\documentclass{article}
\usepackage{amsmath}

\usepackage[main, final]{neurips_2026}

\usepackage[utf8]{inputenc} 
\usepackage[T1]{fontenc}    
\usepackage{hyperref}       
\usepackage{url}            
\usepackage{booktabs}       
\usepackage{amsfonts}       
\usepackage{nicefrac}       
\usepackage{microtype}      
\usepackage{xcolor}         

\usepackage{amssymb}
\usepackage{pifont}
\usepackage{fancyvrb}
\usepackage{fvextra}
\usepackage{makecell}
\usepackage{graphicx} 
\usepackage{xspace}
\usepackage{xcolor}
\usepackage{longtable}
\usepackage{listings}
\usepackage{multirow}
\usepackage{tcolorbox}
\usepackage{xcolor,colortbl}
\title{Rethinking How to Remember: Beyond Atomic Facts in Lifelong LLM Agent Memory}

\vspace{-8mm}
\author{
\textbf{Jingwei Sun}$^{1,\ast}$ \quad \textbf{Jianing Zhu}$^{2,} \thanks{Equal Contribution}$ \quad \textbf{Jiangchao Yao}$^{3}$ \quad \textbf{Tongliang Liu}$^{4}$ \quad \textbf{Bo Han}$^{1,} \thanks{Correspondence to Bo Han (bhanml@comp.hkbu.edu.hk).}$\vspace{1mm}\\
$^{1}$TMLR Group, Hong Kong Baptist University \quad
$^{2}$The University of Texas at Austin \\
$^{3}$CMIC, Shanghai Jiao Tong University \quad
$^{4}$Sydney AI Centre, The University of Sydney \vspace{1mm}\\
\textnormal{\{csjwsun, bhanml\}@comp.hkbu.edu.hk} \quad \textnormal{jianing.zhu@austin.utexas.edu}\\
\textnormal{sunarker@sjtu.edu.cn} \quad \textnormal{tongliang.liu@sydney.edu.au}\\
}

\begin{document}

\maketitle

\begin{abstract}
  To enable reliable long-term interaction, LLM agents require a memory system that can faithfully store, efficiently retrieve, and deeply reason over accumulated dialogue history. Most existing methods adopt an extracted fact based paradigm: handcrafted static prompts compress raw dialogues into atomic facts, which are then stored, matched, and injected into downstream reasoning. Nevertheless, such fact-centric designs inevitably discard fine-grained details in original dialogues and fail to support deep reasoning over scattered isolated facts. Moreover, static prompts cannot maintain consistent extraction granularity across diverse dialogue styles. To address these limitations, we propose TriMem, which maintains three coexisting representation granularities, including raw dialogue segments anchored by source identifiers for storage fidelity, extracted atomic facts for efficient memory retrieval, synthesized profiles that aggregate dispersed facts into holistic semantic understanding for deep reasoning. We further adopt TextGrad-based prompt optimization, which iteratively refines extraction and profiling prompts via response quality feedback, achieving lifelong evolution without any parameter updating. Extensive experiments on LoCoMo and PerLTQA across multiple LLM backbones demonstrate that TriMem consistently outperforms strong memory baselines. The code is publicly available at \url{https://TMLR-TriMem.github.io}.
\end{abstract}

\vspace{-2mm}
\section{Introduction}
The rapid development of large language models (LLMs) has driven significant breakthroughs in agent technology~\cite{huang2024understanding,liu2026frontier,sun2026agenthijack}, which has demonstrated remarkable capabilities in a wide range of scenarios~\cite{sun2026bilateral,qin2025ui,tian2026multi}. However, in practical applications, constrained by the capacity of the context window, text overflow inevitably occurs when it comes to long-context and multi-turn interaction scenarios, leading to significant limitations in agents due to the loss of key historical interaction information and broken context logic~\cite{yan2025memory,yu2025memagent,sheng2026memorize}. To enable reliable long-term interaction, LLM agents require a robust memory system to effectively manage and leverage historical experiences. 

\begin{figure*}[t]
    \centering
    \vspace{-4mm}
    \includegraphics[width=\linewidth]{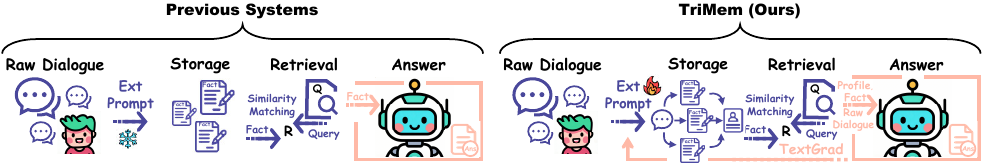}
    \vspace{-2mm}
    \caption{\textbf{Comparison with previous systems.} Our system establishes a three-level architecture, which leverages raw dialogue to guarantee information fidelity in storage, relies on key facts to enable efficient retrieval, and provides in-depth understanding over the facts to ensure the reliability of reasoning. The construction prompts are also continuously optimized based on answer feedback.}
    \label{fig:figure1}
    \vspace{-2mm}
\end{figure*}

Recent research has extensively investigated the design of memory modules for LLM agents~\cite{kang2025memory,chhikara2025mem0,fang2025lightmem,liu2026simplemem,hu2026beyond}, with the overarching goal of realizing three core functional capabilities: storing information from historical interactions, retrieving relevant memories upon query, and incorporating them into prompts to support reasoning. As shown on the left side of Figure~\ref{fig:figure1}, existing systems typically rely on extracted factual information generated by static hand-written prompts to fulfill these three objectives. They only store extracted key facts produced by fixed prompts, perform retrieval based on similarity matching against these facts, and utilize the relevant information to support reasoning, which naturally raises two critical questions: \textit{Can extracted facts beneficially affect all three stages of agent memory systems? and can static fixed prompts adaptively maintain consistent and rational extraction granularity when facing highly heterogeneous real-world dialogue scenarios?}

\begin{figure*}[t]
    \centering
    \includegraphics[width=\linewidth]{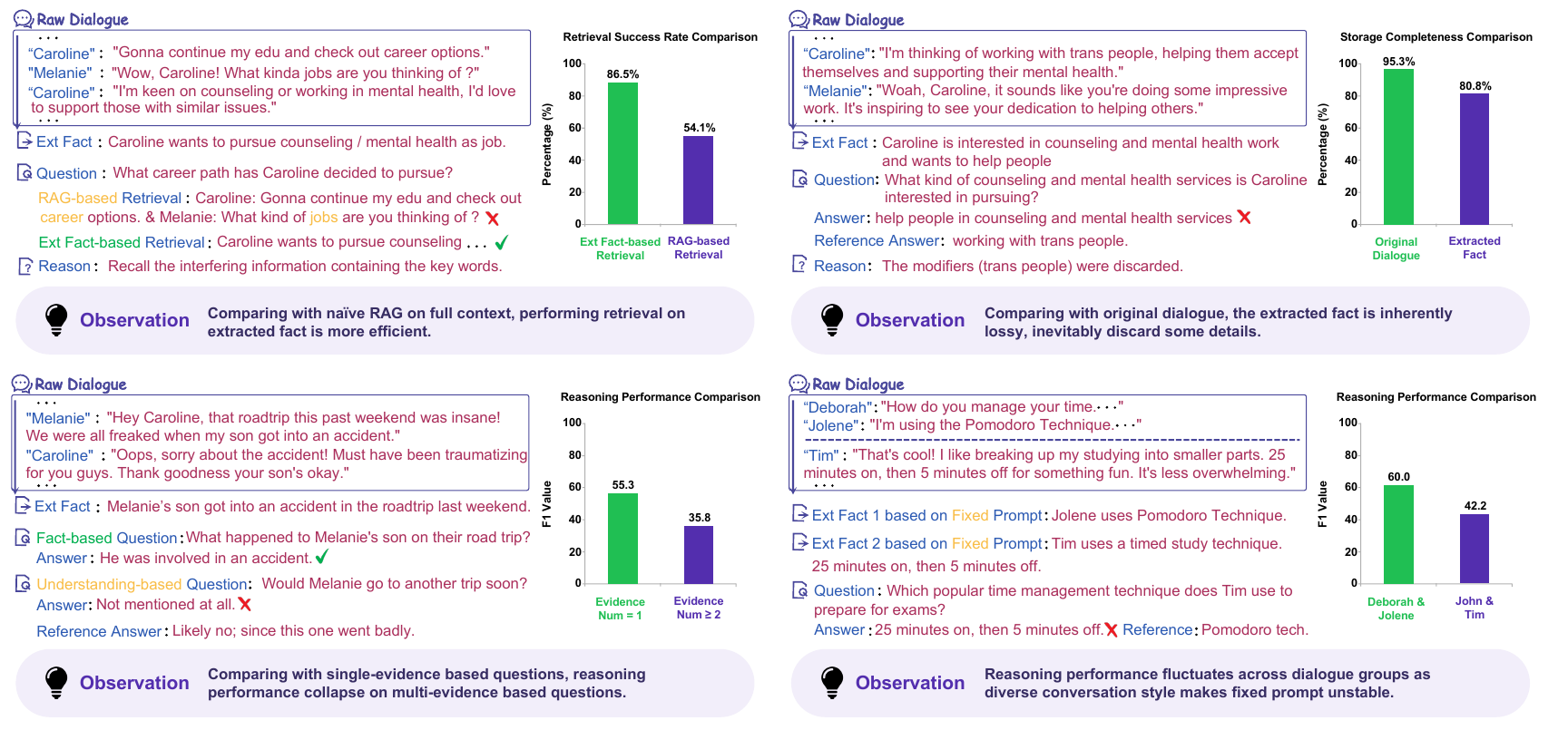}
    \caption{\textbf{Analysis of existing agent memory systems.} Although the totally extracted fact based systems enable efficient retrieval, they suffer from lossy storage and shallow reasoning. Additionally, the fixed prompt words cannot be well applied to all contexts, which compromise the performance.}
    \label{fig:case study of previous method}
    \vspace{-2mm}
\end{figure*}

In this paper, we start by highlighting two concerning phenomena: extracted facts suffer from information loss, and reasoning for many real-world questions relies on in-depth understanding rather than simple fact matching. As shown in Figure~\ref{fig:case study of previous method}, we analyze the performance of existing memory systems from storage fidelity, retrieval efficiency, and reasoning quality. The results demonstrate that although structured factual information ensures efficient retrieval, some precise details are inevitably lost since fact extraction necessarily compresses lengthy dialogues into concise representations. Fortunately, the original dialogue preserves complete information and can serve as an auxiliary source to recover these unstored details. Furthermore, while some questions can be answered using a single retrieved fact, others require integrating scattered information to form a comprehensive understanding. This suggests that a memory system should be capable of further information integration and in-depth comprehension beyond discrete factual pieces. Such a three-level architecture, consisting of raw dialogue storage, extracted fact based retrieval, and understanding-aided reasoning, offers potential for balancing the three core functionalities of a memory system. We also illustrate that the diversity of information styles leads to inconsistent extraction granularity, which significantly degrade system performance. Therefore, we require a mechanism that can continuously optimize prompts, enabling the system to adaptively balance storage depth, retrieval efficiency, and reasoning quality over time.

Based on our findings in the analysis, we propose \textit{\textbf{Tri}-Granularity \textbf{Mem}ory} (TriMem), a memory architecture that maintains three coexisting granularities of representation, including verbatim dialogues, extracted facts, and synthesized profiles. In general, it builds source dialogue identifiers to preserve pointers to original dialogue segments for each compressed memory entry and employs a profile module that incrementally integrates scattered facts during ingestion to form understanding (as shown in the right part of Figure~\ref{fig:figure1}). On the one hand, considering the insufficient retention of detailed information, a traceable link is created for each compressed memory entry, which is directly associated with the original dialogue from which the entry is derived. On the other hand, considering the lack of in-depth understanding of factual information, incremental profile is constructed based on the new information of each window and pre-integrated understanding is provided at reasoning time, thus eliminating the need to perform complex comprehensive analysis among scattered facts. Furthermore, unlike previous methods that utilize hand-written prompts, we employ TextGrad-based prompt optimization~\cite{yuksekgonul2024textgrad} to iteratively refine system prompts, thereby achieving the precise and high-quality construction of the memory module. To demonstrate the effectiveness of our method, we conduct extensive experiments on commonly used benchmarks~\cite{locomo,du2024perltqa} and provide discussions on various ablations to justify the rationality. Our main contributions are summarized as follows:
\begin{itemize}
\item[$\bullet$] Conceptually, we revisit the memory systems of LLM agents from a novel three-dimensional perspective, systematically exploring the limitations of existing paradigms spanning storage fidelity, indexing efficiency and reasoning quality (in Section~\ref{sec:preliminary and motivation}).

\item[$\bullet$] Technically, we propose TriMem, which innovatively introduces entity profile module to avoid shallow reasoning, preserves raw dialogue identifier to avoid detail loss and iteratively optimize system prompts to avoid performance fluctuation (in Section~\ref{sec:method}).
\item[$\bullet$] Experimentally, we conduct extensive explorations to verify the effectiveness of TriMem under different scenarios, including the significant improvement across various benchmarks, compatibility with diverse model structure and model size, etc (in Section~\ref{sec:experiment}).
\end{itemize}

\vspace{-1mm}
\section{Preliminary and Motivation}
\label{sec:preliminary and motivation}
\vspace{-1mm}
In this section, we review the preliminaries of conventional agent memory systems (in Section~\ref{sec:Conventional agent memory systems}), and conduct an in-depth analysis of the inherent limitations of existing agent memory systems (in Section~\ref{sec:Limitation of Existing Systems}). The experimental details of our analysis can be found in Appendix~\ref{app:more experimental details}.

\subsection{Conventional Agent Memory Systems}
\label{sec:Conventional agent memory systems}
We characterize the mainstream memory architecture~\cite{kang2025memory,chhikara2025mem0,liu2026simplemem} into three core phases: \textbf{(I) Storage.} Given a historical dialogue $\mathcal{H}=\left\{\left(r_{t}, u_{t}\right)\right\}_{t=1}^{T}$ consisting of $T$ turns of interaction between the user $r$ and $u$, we first partition it into discrete windows $\mathcal{W}=\{w_1,w_2,...,w_N\}$ following a specific granularity $g\in\{\text{turn},\text{session},\text{topic}\}$. Key factual information is then extracted by one LLM agent $\phi$ and fixed extraction prompt $\mathbf{p}_e$ from each window to generate memory entries $\mathcal{E}=\left\{e_{i}|e_{i}=\phi\left(\mathcal{W}, \mathbf{p}_e \right)\right\}_{i=1}^{K}$. \textbf{(II) Retrieval.} Given question
$q$, relevant entries are retrieved from the memory bank through similarity matching $\mathcal{E}^*(q)=\text{top-}K_{e_i\in\mathcal{E}}\;\text{sim}(e_i,k_q)$, $k_q$ is either the question itself or query about the question, such as keywords. \textbf{(III) Reasoning.} The retrieved entries are concatenated with the question, which is subsequently fed into the language model to generate a final response $a=\phi(\mathcal{E}^*(q),q)$.

\subsection{Analysis of Existing Systems}
\label{sec:Limitation of Existing Systems}
\textbf{Efficient Retrieval.} As illustrated in Figure~\ref{fig:case study of previous method}, we compare one representative system~\cite{liu2026simplemem} with full-context based RAG strategy~\cite{lewis2020retrieval}, in terms of the ratio of successfully retrieved question-related evidence. It can be seen that the extracted fact based retrieval enables more efficient and accurate localization of relevant memory entries. This alleviates the critical limitation of full-context based retrieval, which tends to introduce massive irrelevant and distracting content. For instance, when a query contains the keyword \textit{career}, the full-context based RAG strategy retrieves massive sentences containing lexical variants such as \textit{job} and \textit{career}, where redundant noise severely degrades retrieval efficiency. In contrast, extracted facts are summarized across multi-turn dialogues, thus enabling precise semantic matching and substantially improving retrieval performance. However, the heavy reliance of subsequent phases on these extracted fact introduces several limitations we examine next.

\textbf{Lossy Storage.} Although retrieval on extracted facts enables accurate memory localization, the fact extraction process inevitably introduces lossy compression of raw information, where fine-grained details are discarded, resulting in inherent information incompleteness within stored memory entries. As illustrated in Figure~\ref{fig:case study of previous method}, we calculate the coverage rate of reference answer token, it can be seen that the extracted fact loss 14.5\% more information than original dialogue. For example, the modifier \textit{trans} in the original dialogue is omitted and compressed during fact extraction. This directly leads to a critical issue: even if the system retrieves topically relevant entries, it still fails to provide accurate and complete answers due to missing key contextual details from raw dialogues. This implies that completely abandoning raw dialogues and relying solely on extracted facts for storage will result in permanent loss of semantic details. Such systems become incapable of solving detail-dependent and high-precision constrained queries, thus suffering from inevitable performance degradation.

\textbf{Shallow Reasoning.} Beyond the aforementioned issue of irreversible information loss during extraction, existing extracted fact based memory systems additionally suffer from a fundamental capability bottleneck of shallow reasoning. As illustrated in Figure~\ref{fig:case study of previous method}, we test the reasoning performance of those correctly retrieved questions. The results reveal that the reasoning performance on multi-evidence questions is considerably inferior to that on single-evidence ones, as single-evidence questions mostly rely on explicit factual content and can be answered by directly restating content in retrieved memory entries. In contrast, multi-evidence questions demand deep understanding of dispersed facts, such as emotional inference and logical induction. This phenomenon adequately demonstrates that reasoning mechanisms solely relying on extracted fact completely lack the ability of deep comprehension towards entity semantic portraits or behavioral tendency modeling, thereby severely hindering the comprehensive performance of LLM agents in long-term interactions.

\textbf{Suboptimal Prompt.} In addition to the information loss and shallow reasoning issues incurred by extracted facts, practical real-world deployments further suffer from performance degradation caused by unstable extraction granularity with fixed prompts. Realistic long-term interactions involve highly diverse information styles, expression patterns and content categories, whereas conventional systems rely on static hand-written extraction prompts. Such fixed prompts fail to adaptively accommodate heterogeneous dialogue content, making it impossible to maintain consistent and rational fact extraction granularity. As illustrated in Figure~\ref{fig:case study of previous method}, we compare the reasoning performance across dialogues between different speakers where the performance severely fluctuate. For example, the \textit{Pomodoro technique} is sometimes explicitly mentioned by name, while in other cases it is implicitly described as \textit{25 minutes on and 5 minutes off}. The fixed prompt cannot recognize such high-level semantic concepts, resulting in inconsistent extraction granularity. This discrepancy further degrades the overall performance and stability of the memory system.

\section{Method}
\label{sec:method}
In this section, we present TriMem, which constructs a three-level architecture, from raw dialogue to extracted key factual information and integrated profiles (in Section~\ref{sec:overall pipeline}). Specially, we construct index between raw dialogue and extracted fact to improve storage fidelity while keeping retrieval efficiency (in Section~\ref{sec:memory construction}), integrate scattered facts to profiles to support understanding based reasoning (in Section~\ref{sec:memory retrieval}), optimize prompts via failure case analysis to realize lifelong evolution (in Section~\ref{sec:failure analysis}).

\subsection{Overall Pipeline}
\label{sec:overall pipeline}
\begin{figure*}[t]
    \centering
    \includegraphics[width=\linewidth]{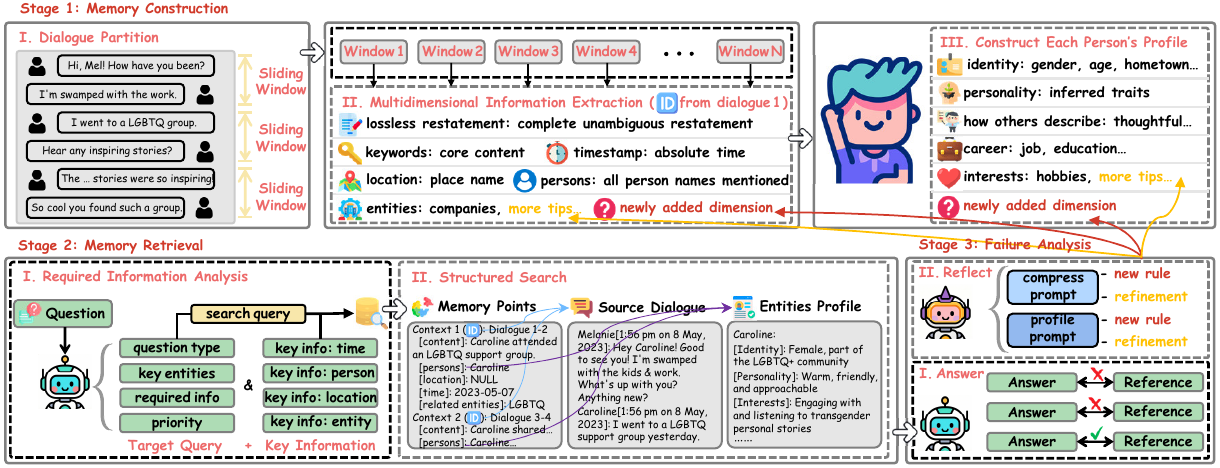}
    \caption{\textbf{Overview of TriMem.} It segments historical dialogue into windows, extracts multi-dimensional facts with traceable index, and constructs entity profiles. Relevant memories are retrieved according to queries, and prompts are continuously optimized via response feedback.}
    \label{fig:pipeline}
    \vspace{-2mm}
\end{figure*}

As shown in Figure~\ref{fig:pipeline}, our framework consists of the following three core components: (1) Given a set of dialogue, we extract factual information and bind each fact with raw dialogue identifier and profile identifier to implement memory construction. (2) Given a question, the agent first analyzes the required information and keywords to generate search query. It then matches with extracted facts to retrieve relevant memory, and further obtains raw dialogues with full details and integrated profiles via predefined index to complete memory retrieval. (3) According to the reasoning performance, we invoke a powerful senior model to conduct failure analysis and provide revision suggestions for system prompts. In the following subsections, we will elaborate on these components in detail.

\subsection{Dense Storage with Efficient Retrieval}
\label{sec:memory construction}
Same as conventional systems, we also perform fact extraction on original dialogue. We first partition the raw dialogue $\mathcal{H}$ into a sequence of overlapping windows via a sliding window $f_{\text{seg}}$:
\begin{equation}
    \mathcal{W} = f_{\text{seg}}(\mathcal{H};\, l,\, s) = \{w_i\}_{i=1}^{N}, \quad
    w_i = \{(r_t, u_t)\}_{t=(i-1)s+1}^{\min(T,\,(i-1)s+l)},
\end{equation}
where $l$ denotes the window size, $s$ denotes the stride, and $N = \lceil(T - l) / s\rceil + 1$ is the total number of windows. The overlapping design ensures that contextually related utterances spanning window boundaries are not artificially severed, thereby preserving local coherence for subsequent extraction. For each window $w_i$, we employ an agent $\phi$ driven by extraction prompt $\mathbf{p}_e$, the extraction prompt is composed of a task instruction and a multi-dimensional schema:
\begin{equation}
    \mathbf{p}_e = \bigl[\,\text{instruction} \;\|\; \text{schema}: \{f_{\text{restate}},\, f_{\text{time}},\, f_{\text{person}},\, \ldots\}\,\bigr],
\end{equation}
where each dimension corresponds to a dedicated extraction function capturing a distinct semantic aspect of the dialogue content: $f_{\text{restate}}$ performs lossless restatement to preserve complete and unambiguous factual content, $f_{\text{time}}$ extracts absolute temporal references, $f_{\text{person}}$ identifies all persons mentioned in the window, and additional dimensions such as location, keywords, and named entities can be flexibly incorporated into the schema. This extensible multi-dimensional design ensures that extraction covers diverse semantic granularities while remaining adaptable to domain-specific requirements. Each extracted entry is thus a structured tuple, \textit{i.e.}, $e_i = \bigl(f_{\text{restate}}^{(i)},\, f_{\text{time}}^{(i)},\, f_{\text{person}}^{(i)},\, \ldots\bigr)$. Critically, unlike previous systems, we add an additional dimension $f_{\text{src}}$ to the extraction prompt to obtain a source dialogue identifier $e_i.\mathrm{src} \triangleq \{r_t, u_t \in w_i\}$ for each entry $e_i$ to address the lossy storage problem. Given question $q$, the system first extracts a set of search query $k_q$ from the question by analysing the required information and relevant keywords. The top-$K$ most relevant entries are then retrieved from $\mathcal{E}$ through similarity matching over extracted fact $\mathcal{E}^*(q) = \operatorname{top-}K_{e_i \in \mathcal{E}}\; \operatorname{sim}(k_q,\, e_i)$. Each retrieved entry $e_i \in \mathcal{E}^*(q)$ can serve as an anchor to recover all original details by the source identifier. In this way, our system guarantees storage fidelity while keeping retrieval efficiency.

\subsection{Scattered Fact with Integrated Profile}
\label{sec:memory retrieval}
To address the shallow reasoning problem, we construct structured profiles over scattered fact. We first group extracted fact according to different person $p \in \mathcal{V}$:
\begin{equation}
\mathcal{E}_p = \{e_i \in \mathcal{E} \mid f_{\text{person}}^{(i)} = p\}, \quad \forall p \in \mathcal{V},
\end{equation}
where $\mathcal{V} = \bigcup_{i=1}^{K} f_{\text{person}}^{(i)}$ denotes the set of all identified persons. Then, driven by a profile prompt $\mathbf{p}_\psi$, the person-specific entries are integrated and synthesized into a structured profile $\mathrm{Prof}(p) = \phi(\mathcal{E}_p,\mathbf{p}_\psi)$, which captures multi-faceted entity-level understanding, including identity attributes (\textit{e.g.}, gender, age, hometown), inferred personality traits, career and educational background, interests and hobbies, as well as interpersonal dynamics and behavioral tendencies. Ultimately, also anchored by the retrieved entry, we can obtain integrated profiles during reasoning through the person identifier. In this way, we can provide agents with understanding over scattered facts to avoid shallow reasoning.


\subsection{Lifelong Evolution with Optimized Prompt}
\label{sec:failure analysis}
In order to maintain the fine-grained consistency, we utilize TextGrad~\cite{yuksekgonul2024textgrad} to optimize the prompts.
Given the retrieved context $\mathcal{R}(q)$, the system generates a response via an answer generation prompt $\mathbf{p}_a$, whose output can be compared against reference answers $a^*$ to obtain a quality signal. We leverage this failure signal to continuously optimize the upstream extraction prompt $\mathbf{p}_e$ and profile prompt $\mathbf{p}_\psi$, treating them as joint trainable parameters $\boldsymbol{\theta} = (\mathbf{p}_e, \mathbf{p}_\psi) \in \Sigma^* \times \Sigma^*$, where $\Sigma^*$ denotes the space of natural language strings. The complete optimization objective is:
\begin{equation}
    \boldsymbol{\theta}^* = \arg\min_{\boldsymbol{\theta}}\, \mathbb{E}_{(q,\, a^*) \sim \mathcal{D}}\bigl[-\mathrm{Score}_{\mathrm{LLM}}\!\left(q,\, \mathrm{LLM}(q,\, \mathcal{R}(q)),\, a^*\right)\bigr].
\end{equation}
where $\mathrm{Score}_{\mathrm{LLM}}$ evaluates answer quality via an LLM judge, compatible with metrics such as F1 and BLEU. Rather than numerical gradients, TextGrad backpropagates the loss through natural language editing suggestions given by LLM agent $\mathrm{TextGrad}_{\mathrm{LLM}}$:
\begin{equation}
    \mathbf{g}_e,\mathbf{g}_\psi = \frac{\partial \mathcal{L}}{\partial \boldsymbol{\theta}}= \mathrm{TextGrad}_{\mathrm{LLM}}\!\left(\mathcal{L},\, \boldsymbol{\theta}\right) \in \Sigma^*.
\end{equation}
The prompts are then updated by applying these textual gradients as rewriting instructions: $\mathbf{p}_e^{(t+1)} = \mathbf{p}_e^{(t)} \oplus \mathbf{g}_e^{(t)}$, $\mathbf{p}_\psi^{(t+1)} = \mathbf{p}_\psi^{(t)} \oplus \mathbf{g}_\psi^{(t)}$, where $\oplus$ denotes the prompt editing operator. Through iterative optimization, $\mathbf{p}_e$ progressively refines extraction granularity to retain semantically valuable details, while $\mathbf{p}_\psi$ improves its capacity to capture behavioral patterns and entity-level understanding, enabling our system to continuously evolve without manual prompt engineering.

\section{Experiments}
\label{sec:experiment}
In this section, we provide comprehensive verification of TriMem. First, we introduce several critical parts of experimental setups (in Section~\ref{sec:exp_setting}). Second, we provide performance comparison and compatibility experiments of Entity Profile with different previous methods (in Section~\ref{sec:main_results}). Third, we conduct extensive ablation studies to better understand our TriMem (in Section~\ref{sec:ablation}).
\subsection{Experimental Setups}
\label{sec:exp_setting}
\textbf{Baselines and Benchmarks.} We compare our method with Naive RAG and several competitive agent memory systems, including Mem0~\cite{chhikara2025mem0}, MemoryOS~\cite{kang2025memory}, A-Mem~\cite{yu2026amem}, LightMem~\cite{fang2025lightmem}, SimpleMem~\cite{liu2026simplemem} and xMemory~\cite{hu2026beyond}. For a fair comparison, we keep the original hyperparameter setups of the comparative methods. The evaluation is conducted on three representative benchmarks, including LoCoMo~\cite{locomo}, PerLTQA~\cite{du2024perltqa} and LongMemEval~\cite{wu2025longmemevalbenchmarkingchatassistants}. More details of each memory system and benchmark, including their evaluation settings and task characteristics, are provided in Appendix~\ref{app:details about baselines and benchmarks}.


\textbf{Implementation Details.} We set the size of the window to 40 and stride to 38. The Qwen3-embedding-0.6b~\cite{zhang2025qwen3embedding} model is utilized to encode the extracted fact. During retrieval, the maximum number of relevant entries is set to 25. The number of optimization rounds for prompts is set to 4 to enable in-depth reasoning. We perform prompt training on a small development subset of LoCoMo, consisting of dialogue0 and dialogue1, using Qwen3-8B~\cite{yang2025qwen3}, and then directly apply the optimized prompts to the remaining dialogues, other models and benchmarks. The failure analysis is conducted by Claude Opus 4.6~\cite{anthropic2026opus46}. The prompts used in our experiments are provided in Appendix~\ref{app:system prompts}.

\subsection{Main Results}
\label{sec:main_results}
\begin{table*}[t]
\centering
\caption{\textbf{Performance on the LoCoMo benchmark with High-Capability Models.} TriMem achieves outstanding performance across different models while maintaining high efficiency.}
\label{tab:high_cap_results}
\resizebox{\textwidth}{!}{
\begin{tabular}{l|l|cc|cc|cc|cc|cc|r}
\toprule
\multirow{2}{*}{\textbf{Model}} & \multirow{2}{*}{\textbf{Method}} & \multicolumn{2}{c|}{\textbf{MultiHop}} & \multicolumn{2}{c|}{\textbf{Temporal}} & \multicolumn{2}{c|}{\textbf{OpenDomain}} & \multicolumn{2}{c|}{\textbf{SingleHop}} & \multicolumn{2}{c|}{\textbf{Average}} & \multicolumn{1}{c}{\textbf{Token}} \\
 & & \textbf{BLEU} & \textbf{F1} & \textbf{BLEU} & \textbf{F1} & \textbf{BLEU} & \textbf{F1} & \textbf{BLEU} & \textbf{F1} & \textbf{BLEU} & \textbf{F1} & \multicolumn{1}{c}{\textbf{Cost}} \\
\midrule

\multirow{8}{*}{\textbf{GPT-4.1-mini}} 
 & LoCoMo & 8.00  & 17.26 & 10.17 & 14.89 & 8.29  & 16.28 & 17.43 & 19.36 & 13.62 & 17.85 & 16863 \\
 & Naïve RAG & 11.49 & 13.24 & 20.52 & 28.80 & 11.79 & 10.75 & 22.85 & 30.29 & 19.59 & 25.64 & 1119  \\
 & Mem0 & 28.81 & 31.44 & 35.41 & 46.24 & 18.51 & 17.93 & 31.25 & 35.34 & 30.88 & 35.81 & 1153  \\
 & MemoryOS & 16.46 & 24.02 & 34.78 & 46.52 & 14.89 & 19.58 & 36.18 & 43.92 & 30.95 & 39.30 & 936 \\
 & A-Mem & 15.11 & 20.66 & 41.57 & 50.94 & 11.18 & 13.20 & 38.25 & 43.72 & 33.01 & 39.10 & 1276  \\
 & LightMem & 32.93 & 40.33 & 47.53 & 55.23 & 18.31 & 21.91 & 37.68 & 48.39 & 37.66 & 46.69 & 695   \\
 & SimpleMem & 32.40 & 39.33 & 43.69 & 58.01 & 19.56 & 24.50 & 43.41 & 53.99 & 39.97 & 50.30 & 587 \\
 \rowcolor{gray!9} \cellcolor{white}& \cellcolor{white}\textbf{Ours} &  \textbf{35.20} & \textbf{42.59} & \textbf{49.56} & \textbf{64.72} & \textbf{36.86} & \textbf{43.88} & \textbf{45.25} & \textbf{55.36} & \textbf{43.79} & \textbf{54.26} & 1217 \\

\midrule

\multirow{8}{*}{\textbf{GPT-4o}} 
  & LoCoMo & 19.64 & 19.20 & 9.50  & 13.95 & 11.87 & 16.60 & 13.81 & 16.12 & 13.86 & 16.26 & 16863 \\
 & Naïve RAG & 14.36 & 15.35 & 11.48 & 16.17 & 9.03  & 9.09  & 26.67 & 35.03 & 20.15 & 25.88 & 1119  \\
 & Mem0 & 25.52 & 32.36 & 32.48 & 42.70 & 14.50 & 18.50 & 30.02 & 39.84 & 28.74 & 37.74 & 1195  \\
 & MemoryOS & 22.52 & 31.76 & 38.31 & 47.08 & 12.91 & 18.06 & 38.26 & 43.67 & 33.81 & 40.60 & 944   \\
 & A-Mem & 20.90 & 26.12 & 35.39 & 48.64 & 10.74 & 12.33 & 37.11 & 42.08 & 32.14 & 38.67 & 1152  \\
 & LightMem & 35.30 & 45.16 & 43.60 & 58.57 & 10.56 & 23.20 & 36.72 & 46.60 & 36.26 & 47.37 & 677   \\
 & SimpleMem & 31.34 & 35.58 & 35.78 & 46.96 & 18.96 & 17.01 & 37.11 & 43.94 & 34.64 & 41.36 & 627   \\
 \rowcolor{gray!9} \cellcolor{white} & \cellcolor{white}\textbf{Ours} & \textbf{40.36} & \textbf{46.00} & \textbf{51.39} & \textbf{60.41} & \textbf{39.27} & \textbf{50.15} & \textbf{40.61} & \textbf{47.78} & \textbf{42.73} & \textbf{50.23} & 1272 \\
\midrule

\multirow{8}{*}{\textbf{GPT-5-nano}} 
 & LoCoMo & 20.45 & 19.04 & 12.69 & 16.56 & 13.83 & 20.85 & 13.50 & 15.23 & 14.62 & 16.56 & 16863 \\
 & Naïve RAG & 10.13 & 13.29 & 8.78  & 13.09 & 9.25  & 12.24 & 20.29 & 28.44 & 15.34 & 21.46 & 1119  \\
 & Mem0 & 22.55 & 28.58 & 35.52 & 48.82 & 18.33 & 16.75 & 28.99 & 35.65 & 28.51 & 35.92 & 1074  \\
 & MemoryOS & 10.74 & 23.50 & 32.50 & 39.71 & 10.02 & 20.30 & 34.28 & 40.34 & 28.09 & 35.88 & 952   \\
 & A-Mem & 15.54 & 20.11 & 27.23 & 32.43 & 10.86 & 12.55 & 27.26 & 31.91 & 24.09 & 28.65 & 1175  \\
 & LightMem & 28.63 & 38.21 & 39.72 & 55.51 & 18.79 & 22.74 & 31.19 & 42.01 & 31.73 & 42.93 & 723   \\
 & SimpleMem & 25.42 & 33.28 & 32.15 & 45.75 & 20.77 & 24.31 & 39.65 & 46.71 & 34.30 & 42.65 & 655   \\
 \rowcolor{gray!9} \cellcolor{white} & \cellcolor{white}\textbf{Ours} & \textbf{34.86} & \textbf{45.25} & \textbf{42.45} & \textbf{57.05} & \textbf{33.55} & \textbf{40.52} & \textbf{54.26} & \textbf{62.88} & \textbf{46.96} & \textbf{57.04} & 1256\\
\bottomrule
\end{tabular}
}
\vspace{-2mm}
\end{table*}

\begin{table*}[t]
\centering
\caption{\textbf{Performance on the LoCoMo benchmark with Efficient Models.} TriMem still achieves superior performance-efficiency balance with representative efficient models.}
\label{tab:low_cap_results}
\resizebox{\textwidth}{!}{
\begin{tabular}{l|l|cc|cc|cc|cc|cc|r}
\toprule
\multirow{2}{*}{\textbf{Model}} & \multirow{2}{*}{\textbf{Method}} & \multicolumn{2}{c|}{\textbf{MultiHop}} & \multicolumn{2}{c|}{\textbf{Temporal}} & \multicolumn{2}{c|}{\textbf{OpenDomain}} & \multicolumn{2}{c|}{\textbf{SingleHop}} & \multicolumn{2}{c|}{\textbf{Average}} & \multicolumn{1}{c}{\textbf{Token}} \\
 & & \textbf{BLEU} & \textbf{F1} & \textbf{BLEU} & \textbf{F1} & \textbf{BLEU} & \textbf{F1} & \textbf{BLEU} & \textbf{F1} & \textbf{BLEU} & \textbf{F1}  & \multicolumn{1}{c}{\textbf{Cost}} \\
\midrule

\multirow{8}{*}{\textbf{Qwen3-8B}} 
 & LoCoMo & 12.67 & 20.54 & 12.32 & 18.55 & 10.59 & 14.39 & 19.76 & 23.78 & 16.34 & 21.51 & 16863 \\
 & Naïve RAG &  13.75 & 15.83 & 8.86  & 13.53 & 9.68  & 10.77 & 21.57 & 28.76 & 16.75 & 22.10 & 1119  \\
 & Mem0 &  28.32 & 30.07 & 23.15 & 26.15 & 11.79 & 15.15 & 30.75 & 34.97 & 27.54 & 30.10 & 1140  \\
 & MemoryOS & 14.38 & 22.72 & 18.67 & 22.79 & 11.06 & 13.52 & 25.65 & 33.52 & 21.22 & 28.06 & 911   \\
 & A-Mem & 16.02 & 21.08 & 28.10 & 37.51 & 14.01 & 14.19 & 33.60 & 40.77 & 28.01 & 34.83 & 1180  \\
 & LightMem & 22.84 & 32.54 & 37.62 & 48.37 & 18.05 & 19.02 & 23.03 & 31.37 & 25.73 & 34.36 & 740   \\
 & SimpleMem & 23.39 & 30.39 & 24.66 & 34.51 & 14.04 & 15.39 & 35.73 & 41.26 & 29.81 & 36.25 & 608   \\
 & xMemory & 28.44 & 39.13 & 28.65 & 35.41 & 17.76 & 21.57 & 40.66 & 50.57 & 34.49 & 43.51 & 2230  \\

 \rowcolor{gray!9} \cellcolor{white} & \cellcolor{white}\textbf{Ours} & \textbf{33.09} & \textbf{41.22} & \textbf{38.71} & \textbf{53.13} & \textbf{30.59} & \textbf{37.64} & \textbf{45.10} & \textbf{52.52} & \textbf{40.66} & \textbf{49.65} & 1339 \\
\midrule

\multirow{8}{*}{\textbf{Llama-3.1-8B-Ins}} 
 & LoCoMo & 13.73 & 23.36 & 13.15 & 20.30 & 11.54 & 19.42 & 18.64 & 25.86 & 16.15 & 23.84 & 16863\\
 & Naïve RAG & 14.31 & 16.18 & 7.63 & 12.28 & 9.09 & 11.58 & 22.99 & 31.02 & 17.33 & 23.18 & 1119\\
 & Mem0 & 13.27 & 16.40 & 8.26 & 12.62 & 7.45 & 8.45 & 21.75 & 31.28 & 16.49 & 23.24 & 1085\\
 & MemoryOS & 13.57 & 22.63 & 19.18 & 23.31 & 10.59 & 13.01 & 23.46 & 31.05 & 19.95 & 26.77 & 964\\
 & A-Mem & 15.80 & 22.84 & 23.79 & \textbf{36.19} & 11.19 & 12.51 & 31.19 & 37.86 & 25.58 & 33.18 & 1340\\
 & LightMem & 13.19 & 19.64 & 16.93 & 28.06 & 17.39 & 20.68 & 27.62 & 41.06 & 22.11 & 33.16 & 758
\\
 & SimpleMem & 18.81 & 26.22 & 21.15 & 30.44 & 15.81 & 18.77 & 26.79 & 31.23 & 23.47 & 29.37 & 674\\
 & xMemory & 21.89 & 31.24 & 21.78 & 26.84 & 12.37 & 16.62 & 27.75 & 41.36 & 24.47 & 34.94 & 2375\\
 \rowcolor{gray!9} \cellcolor{white} & \cellcolor{white} \textbf{Ours} & \textbf{25.42} & \textbf{34.56} & \textbf{25.98} & 32.36 & \textbf{28.40} & \textbf{32.71} & \textbf{35.76} & \textbf{43.20} & \textbf{31.37} & \textbf{38.70} & 1388
\\
\bottomrule
\end{tabular}
}
\vspace{-2mm}
\end{table*}

\begin{table*}[t]
\centering
\caption{Performance comparison on the PerLTQA benchmark. All values report the LLM-judged correctness score (\%), where \texttt{gpt-4.1-mini} evaluates whether each predicted answer matches the ground truth. TriMem achieves the best performance across different sub-tasks.}
\label{tab:longmemeval_full}
\resizebox{\textwidth}{!}{%
\begin{tabular}{l|cccc|cccc}
\toprule
\multirow{2}{*}{\textbf{Method}} & \multicolumn{4}{c|}{\textbf{Qwen3-8B}} & \multicolumn{4}{c}{\textbf{Llama-3.1-8B-Ins}}\\
 & \textbf{Profile} & \textbf{Social Relationship} & \textbf{Events} & \textbf{Dialogues} & \textbf{Profile} & \textbf{Social Relationship} & \textbf{Events} & \textbf{Dialogues} \\
\midrule
 Full-Context & 65.80 & 56.72 & 52.75 & 18.51 & 52.46 & 54.58 & 47.54 & 17.27 \\
 Mem0 & 89.56 & 76.46 & 66.48 & 27.59 & 73.04 & 72.29 & 57.14 & 26.31 \\
 LightMem & 64.93 & 78.00 & 73.03 & 47.01 & 53.85 & 74.08 & 69.21 & 44.72 \\
 SimpleMem & 88.12 & 82.40 & 79.87 & 42.09 & 84.64 & 76.46 & 70.39 & 37.90 \\
 \rowcolor{gray!9} \cellcolor{white}\textbf{TriMem} & \textbf{92.46} & \textbf{83.23} & \textbf{85.72} & \textbf{55.79} & \textbf{92.17} & \textbf{82.28} & \textbf{78.17} & \textbf{45.01}\\
\bottomrule
\end{tabular}
}
\vspace{-1em}
\end{table*}

\begin{table*}[t]
\centering
\caption{\textbf{Performance comparison on the LongMemEval benchmark.} We report the LLM-judged correctness score of different memory systems with GPT-4.1-mini. TriMem achieves the best overall performance while exhibiting consistently competitive results across different sub-tasks.}
\label{tab:longmemeval_full_2}
\resizebox{\textwidth}{!}{%
\begin{tabular}{l|ccccccc}
\toprule
\textbf{Method} & \textbf{Temporal}	& \textbf{Multi-Session} & \textbf{Knowledge-Update} & \textbf{Single-Session-User} & \textbf{Single-Session-Assistant} & \textbf{Single-Session-Preference} & \textbf{Average}\\
\midrule
Full-context & 27.06\% & 30.08\% & 41.03\% & 47.14\% & 32.14\% & 60.00\% & 39.57\%\\
Mem0 & 40.60\% & 50.37\% & 69.23\% & 87.14\% & 48.21\% & 63.33\% & 59.81\%\\
LightMem & \textbf{85.71\%} & 47.37\% & \textbf{92.30\%} & 88.57\% & 21.43\% & 76.67\% & 68.67\%\\
SimpleMem & 83.46\% & 60.92\% & 79.48\% & 85.71\% & 75.00\% & 76.67\% & 76.87\% \\
\rowcolor{gray!9} \cellcolor{white} \textbf{TriMem} & 79.70\% & \textbf{72.18\%} & 91.03\% & \textbf{94.29\%} & \textbf{91.07\%} & \textbf{80.00\%} & \textbf{82.80\%} \\
\bottomrule
\end{tabular}
}
\vspace{-2mm}
\end{table*}

\textbf{Performance on High-Capability Models.}
In Table~\ref{tab:high_cap_results}, we compare representative agent memory systems to validate the effectiveness of TriMem. Results demonstrate that our method consistently outperforms prior approaches when integrated with various high-capability models, including GPT-4o~\cite{hurst2024gpt4o}, GPT-4.1-mini~\cite{achiam2023gpt4} and GPT-5-nano~\cite{singh2025gpt5}. Meanwhile, to illustrate the information density of TriMem, we report the average tokens consumption of retrieved contexts. It can be seen that our system only consumes around 1.2k tokens. Although retrieving raw dialogues and entity profiles slightly increases the token overhead, it brings substantial performance gains.

\textbf{Compatibility with Efficient Models.} To evaluate the capability of TriMem to support small-parameter efficient models, we conduct experiments on lightweight models including Qwen3-8B~\cite{yang2025qwen3} and Llama-3.1-8B-Instruct~\cite{grattafiori2024llama3}. As shown in Table~\ref{tab:low_cap_results}, our system still achieves substantial performance improvements. Unlike xMemory~\cite{hu2026beyond}, which only supports limited open-source models due to its requirement for model output logits, TriMem is compatible with models of various parameter sizes. This verifies the extensive superiority and broad applicability of our method.

\textbf{Generalization on Different Datasets.}
PerLTQA and LongMemEval are also widely adopted benchmarks for long-term agent QA. PerLTQA consists of multi-dimensional evaluations covering personal profiles, social relationship, historical events and dialogue memories. LongMemEval evaluates five core memory abilities, including information extraction, multi-session reasoning, knowledge updates, temporal reasoning, and abstention. To comprehensively validate the effectiveness of our proposed method, we additionally report experimental results on these two datasets in Table~\ref{tab:longmemeval_full} and Table~\ref{tab:longmemeval_full_2}, which demonstrates the strong generalization ability of TriMem.
\subsection{Ablation and Further Analysis}
\label{sec:ablation}
\begin{figure*}[!t]
    \centering
    \includegraphics[width=\linewidth]{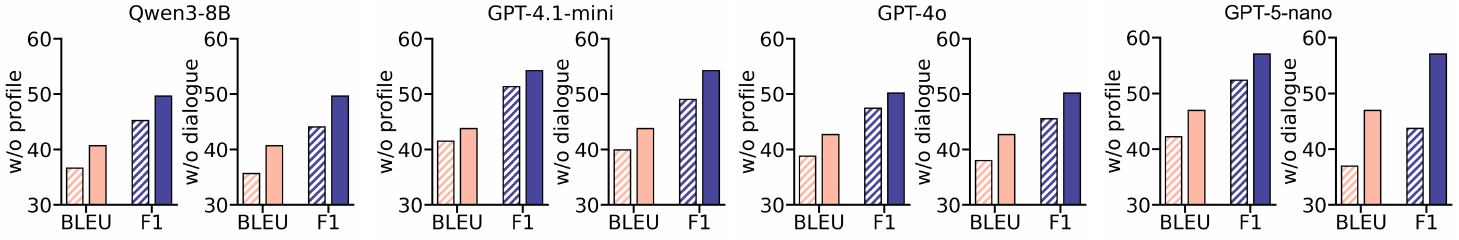}
    \vspace{-6mm}
    \caption{\textbf{Ablation of profile and raw dialogue module.} Incorporating entity profile and raw dialogue shows the best results, which demonstrates the rationality of our design.}
    \label{fig:ablation_profile_dialogue}
    \vspace{-2mm}
\end{figure*}

\textbf{Ablation of Profile and Raw Dialogue.} To verify the necessity of entity profiles and raw dialogues for reasoning reliability and storage fidelity, we conduct ablation experiments on LoCoMo by removing these two contextual components. The results are presented in Figure~\ref{fig:ablation_profile_dialogue}, it can be seen that the agent suffers a noticeable performance drop when either component is excluded. This demonstrates that incorporating entity profiles and raw dialogues is a reasonable design which effectively boosts the overall capability of agent memory systems. More detailed results can be found in Appendix~\ref{app:ablation of profile and raw dialogue}.

\begin{figure*}[!t]
    \centering
    \includegraphics[width=\linewidth]{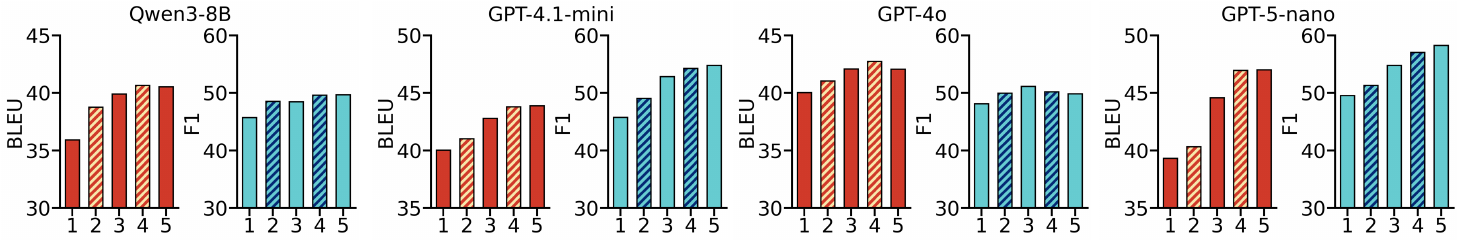}
    \vspace{-6mm}
    \caption{\textbf{Impact of varying evolution step.} Appropriate update steps lead to overall improvements, while further updates result in excessive refinement, which negatively affects the performance.}
    \label{fig:ablation_step}
    \vspace{-2mm}
\end{figure*}

\textbf{Impact of Varying Evolution Step.} We perform multi-step iterative evolution of prompt. To evaluate the impact of different evolution steps on overall system performance, we conduct an ablation study by increasing the number of update steps from 1 to 5. As shown in Figure~\ref{fig:ablation_step}, the performance gradually improves with the increase of evolution steps, and reaches the optimal level when the number of steps is set to 4. Further updates lead to excessive refinement of the prompt’s granularity, which may exceed the model’s capability boundary and result in performance degradation. Therefore, we set the number of evolution steps to 4. Visualization of update process can be found in Appendix~\ref{app:visualization of prompt evolution}.

\begin{figure*}[!t]
    \centering
    \includegraphics[width=\linewidth]{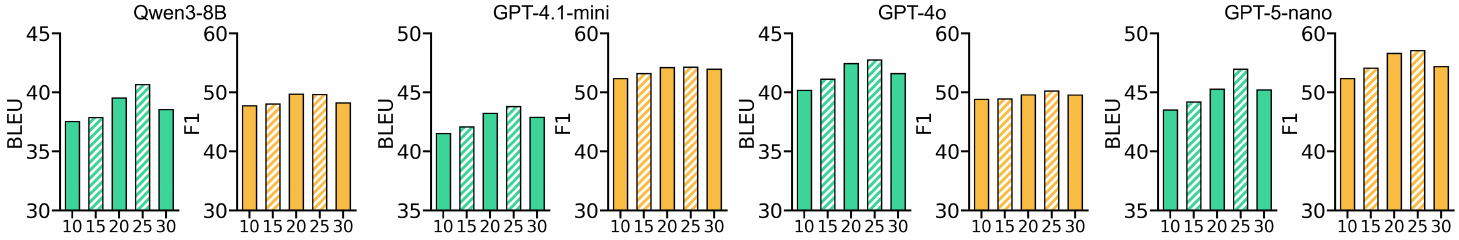}
    \vspace{-6mm}
    \caption{\textbf{Performance in different retrieval numbers.} The results show that the performance is optimal when the number is 25, and performance degrades when the number is too low or too high.}
    \label{fig:ablation_topk}
    \vspace{-2mm}
\end{figure*}

\textbf{Performance in Different Retrieval Numbers.} In Figure~\ref{fig:ablation_topk}, we present the performance of our method when retrieving different numbers of memory entries. It can be observed that when the number of retrieved memory entries is too low, key information may fail to be effectively retrieved, resulting in suboptimal system performance. When the number of retrieved entries is excessively high, irrelevant redundant information is likely to be introduced, interfering with model reasoning and leading to performance degradation. Overall, setting the number of retrieved memory entries to 25 is a reasonable choice. Visualization of the retrieved memory entries can be found in Appendix~\ref{app:content of retrieval entries}.

\begin{figure*}[t]
    \centering
    \includegraphics[width=\linewidth]{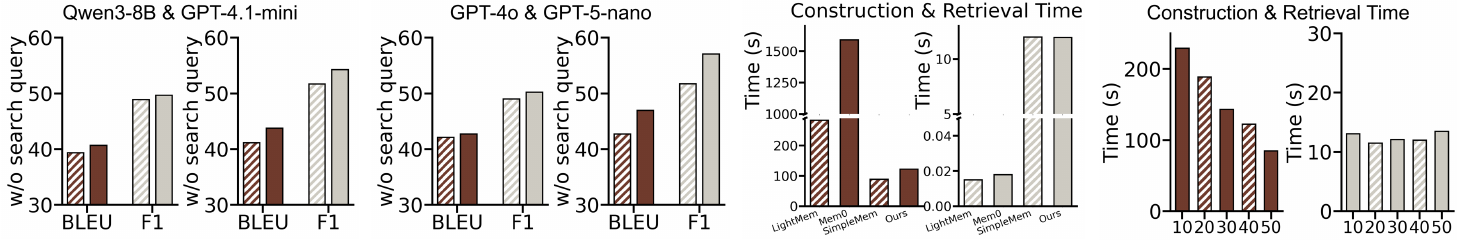}
    \vspace{-6mm}
    \caption{\textbf{Necessity of search query and efficiency analysis.} Although the introduction of search query increases the retrieval time, it greatly improve the system performance. The smaller window size increase the memory construction time, therefore we set the size to 40 to maintain the efficiency, which enabling competitive construction time comparing with previous methods.}
    \label{fig:ablation_window_time}
    \vspace{-2mm}
\end{figure*}

\textbf{Necessity of Search Query.} In the retrieval phase, instead of directly relying on the original question for retrieval, we prompt the agent to first analyze the required information and keywords of the question, and then perform retrieval based on the analysis results. To verify the necessity of this module, we compare the system performance with and without search query in Figure~\ref{fig:ablation_window_time}. The experimental results show that although the generation process of search query increase the retrieval time, the performance is significantly improved after adding the search query, indicating that the search query helps the model achieve more accurate retrieval, thus confirming the necessity of this module. We show some obtained search queries of different questions in Appendix~\ref{app:example of search query}.

\begin{figure*}[t]
    \centering
    \includegraphics[width=\linewidth]{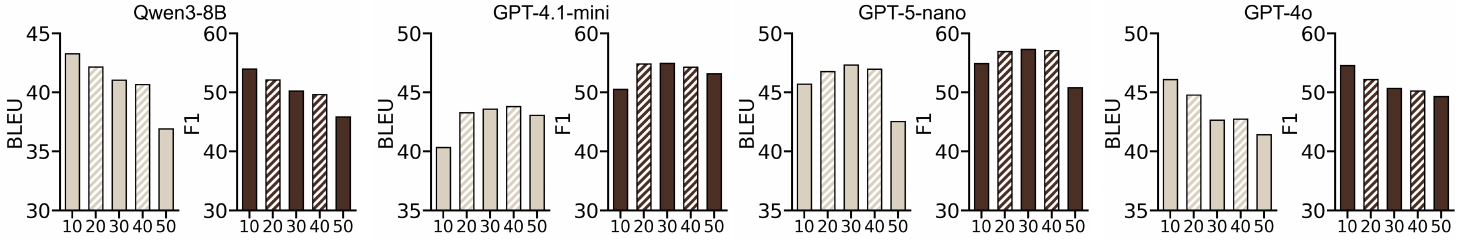}
    \vspace{-6mm}
    \caption{\textbf{Ablation of window size.} The memory system can also achieve great performance with smaller size. However, considering the efficiency, we finally set the window size to 40.}
    \label{fig:ablation_window_size}
    \vspace{-2mm}
\end{figure*}

\textbf{Comparison with Diverse Window Size.} To compare the system performance under different window sizes, we present the performance comparison with various window settings in Figure~\ref{fig:ablation_window_size}. It can be observed that, although the model can also achieve good performance with smaller windows, the number of communications with the agent significantly increases. As shown on the right side of Figure~\ref{fig:ablation_window_time}, although the inference time is not affected, the memory construction time highly extends. Therefore, to balance system performance and efficiency, we finally set the window size to 40. Examples of information extraction under different window sizes can be found in Appendix~\ref{app:performance of different window size}.

\begin{figure*}[!t]
    \centering
    \includegraphics[width=\linewidth]{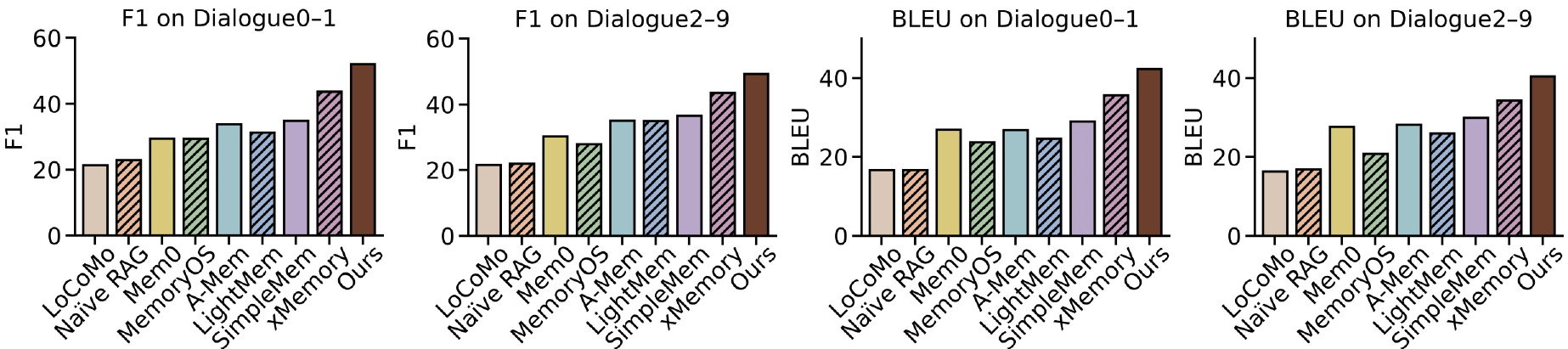}
    \vspace{-6mm}
    \caption{\textbf{Details on two subsets.} The memory system can consistently achieve better performance on both subsets, indicating the stability and generalization ability of our prompt evolution method.}
    \label{fig:performance_on_subset}
    \vspace{-2mm}
\end{figure*}

\textbf{Detailed Results on Two Subsets of LoCoMo.} As illustrated in Section~\ref{sec:exp_setting}, prompt evolution was conducted exclusively on dialogue0 and dialogue1 of LoCoMo using Qwen3-8B. To evaluate the generalizability of the selected prompts, we report TriMem's performance with Qwen3-8B on two LoCoMo subsets in Figure~\ref{fig:performance_on_subset}. TriMem consistently outperforms the baselines on both subsets, indicating that prompt optimization does not overfit to the two development dialogues.

\section{Related Work}
\textbf{Memory Systems for LLM Agents.}
Long-term memory has emerged as a key capability for LLM agents engaged in multi-session interactions, with most existing systems organized around three functional stages: memory construction, memory retrieval, and memory-supported reasoning. For memory construction, Mem0~\citep{chhikara2025mem0}, A-Mem~\citep{yu2026amem}, and MemoryOS~\citep{kang2025memory} prompt LLMs to extract atomic facts from dialogues and consolidate them via dynamic updates or hierarchical stores. For memory retrieval, methods from Naive RAG~\citep{lewis2020retrieval} to xMemory~\citep{hu2026beyond} match queries with fact embeddings through similarity search. For memory-supported reasoning, retrieved facts are concatenated into the prompt as contextual supplements~\citep{liu2026simplemem,fang2025lightmem}. Despite their differences, these systems share a common design choice: extracted facts serve as the atomic unit across all three stages, \textit{i.e.}, what is stored, what is matched, and what is injected into the prompt. This fact-centric pipeline risks losing fidelity to the original dialogue and limits the agent's ability to handle queries that require holistic understanding rather than discrete fact lookup. In contrast, our work revisits the role of each stage, establishing index between extracted facts and source dialogues to ensure storage fidelity and constructing entity profiles from these facts to support comprehension-oriented reasoning that goes beyond fact recall. This perspective also clarifies why the three stages should not be designed in isolation: the granularity of what is stored constrains what can be retrieved,
and the granularity of what is retrieved in turn dictates the kind of reasoning the agent can perform downstream. Treating the three stages as a single pipeline with consistent yet multi-level representations, rather than three independently optimized modules over the same atomic unit, is therefore central to our design.

\textbf{Lifelong Evolution Agents.}
Lifelong evolution has emerged as a key capability for LLM agents that need to improve from accumulated experience. One prominent line formulates this as a reinforcement learning problem: MemAgent~\citep{yu2025memagent} reshapes long-context LLMs into multi-conversation memory agents, MemBuilder~\citep{shen2026membuilder} reinforces memory construction with attributed dense rewards, AgentFold~\citep{ye2025agentfold} learns proactive context management for long-horizon web agents, MemGen~\citep{zhang2025memgen} weaves generative latent memories into self-evolving agents, and MEM1~\citep{zhou2025mem1} jointly synergizes memory and reasoning for efficient execution. While effective, these methods all require RL-based parameter updates that disrupt the original pretrained weights, incur substantial training cost, and are impractical when only API-accessible models are available. Motivated by these limitations, a complementary line pursues lifelong evolution without altering model parameters: Voyager~\citep{wang2023voyager} maintains a skill library of executable code, ExpeL~\citep{zhao2024expel} distills cross-task experience into natural-language rules, and MemSkill~\citep{zhang2026memskill} learns task-agnostic skill memories that transfer to unseen tasks. In line with this paradigm, our work applies TextGrad~\citep{yuksekgonul2024textgrad} to evolve the fact extraction and entity profile construction prompts, achieving lifelong adaptability without tuning parameters. Compared with skill or rule based externalizations that target the agent's action space, prompt-level evolution directly reshapes how raw experience is parsed into memory, which is particularly suited to memory systems whose behavior is largely determined by the extraction and synthesis prompts rather than by a fixed policy.

\section{Conclusion}
In this work, we revisit the design of memory systems for LLM agents and identify three concrete limitations of the prevailing extracted fact based paradigm: lossy storage, shallow reasoning, and suboptimal prompts under heterogeneous dialogue styles. Motivated by these findings, we propose TriMem, which maintains three coexisting representation granularities, including verbatim dialogues for storage fidelity, atomic facts for retrieval efficiency and progressively synthesized profiles for deep reasoning. TextGrad-based prompt optimization is further employed to enable lifelong evolution. Extensive experiments on LoCoMo and PerLTQA across various LLMs confirm the effectiveness of our method. We hope TriMem can bring new insights for future researches on agent memory system.

\section*{Acknowledgments}
JWS and BH were supported by RGC Research Fellow Scheme No. RFS2627-2S02, RGC Young Collaborative Research Grant No. C2005-24Y, RGC General Research Fund No. 12202026, RGC General Research Fund No. 12200725, NSFC Major Research Plan No. 92570109, NSFC General Program No. 62376235, Guangdong and Hong Kong Universities “1+1+1” Joint Research Collaboration Scheme, HKBU Faculty Niche Research Areas No. RC-FNRA-IG/25-26/SCI/05, and HKBU CSD Departmental Incentive Scheme. JCY is supported by National Natural Science Foundation of China (No. 62676245), the Lingang Laboratory (Grant No. LGL-8889-01), STCSM (No. 22DZ2229005) and 111 plan (No. BP0719010). TLL was partially supported by the following Australian Research Council projects: DP260102466, FT220100318, DP220102121, LP220100527, LP220200949.
\newpage

\bibliography{neurips_2026}
\bibliographystyle{plain}

\newpage
\appendix

\section*{Reproducibility Statement}
To ensure the reproducibility of the experimental results, we provide the source code on this link: \href{https://TMLR-TriMem.github.io}{https://TMLR-TriMem.github.io}. In the following, we summarize some critical factors that facilitate reproduction.
\begin{itemize}
\item[$\bullet$]\textbf{Datasets.} The datasets we used are all publicly accessible, which is introduced in Section~\ref{sec:exp_setting}. Following previous work, we test four QA categories on LoCoMo, including MultiHop, Temporal, OpenDomain and SingleHop.
\item[$\bullet$]\textbf{Assumption.} We set up our experiments to agent memory scenario, where a well-pretrained open-source or close-source large language model is available. At every moment, the agent receives a piece of information. It needs to store it in the memory module to assist with potential future reasoning. During the memory process, the agent cannot obtain the relevant information of the problem, it can only rely on the analysis of the original information for memory storage.
\item[$\bullet$]\textbf{Environment.} All experiments are conducted on NVIDIA 4090-24GB GPUs with Python 3.10.
\end{itemize}
\section{Details about Baselines and Benchmarks}
\label{app:details about baselines and benchmarks}
In this section, we provide details about the baselines and corresponding hyperparameters, as well as other related benchmarks that are used in our work.

\paragraph{Full-Context (LoCoMo)~\cite{locomo}.}
The entire conversation history is concatenated and fed directly into the LLM's context window as a single long prompt, without any compression or retrieval. This approach serves as an accuracy upper bound since the model has access to all available information, but it incurs prohibitively high token costs and latency that scale linearly with conversation length, rendering it impractical for real-world long-horizon deployments.

\paragraph{Naive RAG~\cite{lewis2020retrieval}.}
All dialogue turns are chunked into fixed-size segments, embedded using a text encoder, and indexed in a vector store. At inference time, the top-$k$ most similar chunks to the current query are retrieved by cosine similarity and concatenated into the prompt for answer generation. No memory consolidation, update, or deletion is performed; the retrieved context is the raw stored text.

\paragraph{Mem0~\cite{chhikara2025mem0}.}
Mem0 is a scalable memory-centric architecture that dynamically extracts, consolidates, and retrieves salient information from ongoing multi-session conversations. For each new message pair, an LLM extraction function identifies candidate facts, which are then compared against existing memories to determine whether to ADD, UPDATE, DELETE, or NOOP. Retrieved memories are prepended to the prompt at inference time. An enhanced variant, $\text{Mem0}^g$, further represents memories as nodes in a directed graph to capture relational structures among conversational elements.

\paragraph{MemoryOS~\cite{kang2025memory}.}
MemoryOS draws inspiration from operating system memory management to propose a hierarchical storage architecture for AI agents, comprising short-term memory (STM), mid-term memory (MTM), and long-term personal memory (LPM). STM-to-MTM updates follow a dialogue-chain-based FIFO principle, while MTM-to-LPM promotion uses a segmented page organization strategy with a heat-score mechanism that balances visit frequency and recency. A persona module captures evolving user preferences through personalized trait extraction.

\paragraph{A-Mem~\cite{yu2026amem}.}
A-Mem is an agentic memory system inspired by the Zettelkasten method, which organizes memories as dynamically interlinked notes rather than fixed schema. When a new memory is added, the system generates a comprehensive note with structured attributes—contextual descriptions, keywords, and tags—then analyzes the historical memory repository to establish semantic links with relevant prior notes. A memory evolution mechanism allows newly integrated memories to trigger updates to the contextual representations of existing notes, enabling the memory network to continuously refine its organization.

\paragraph{LightMem~\cite{fang2025lightmem}.}
LightMem is an efficiency-oriented memory system modeled after the Atkinson-Shiffrin human memory model. It organizes memory into three complementary stages: a cognition-inspired sensory memory that filters irrelevant content via lightweight compression and groups it by topic; a topic-aware short-term memory that consolidates topic-based groups into structured summaries; and a long-term memory with sleep-time update that decouples consolidation from online inference through an asynchronous offline process. LightMem is designed to dramatically reduce token usage and API calls while maintaining competitive QA accuracy.

\paragraph{SimpleMem~\cite{liu2026simplemem}.}
SimpleMem is an efficient lifelong memory framework built on semantic lossless compression. It adopts a three-stage pipeline: (1) Semantic Structured Compression, which applies entropy-aware filtering to distill raw dialogues into compact, multi-view indexed memory units with resolved coreferences and absolute timestamps; (2) Online Semantic Synthesis, an intra-session process that integrates related units into higher-level abstract representations to eliminate redundancy; and (3) Intent-Aware Retrieval Planning, which infers search intent to dynamically determine retrieval scope and construct precise context efficiently. Memory is indexed using dense semantic embeddings, sparse BM25, and SQL-based metadata storage.

\paragraph{xMemory~\cite{hu2026beyond}.}
xMemory addresses the mismatch between standard RAG assumptions and the properties of agent memory, which forms a bounded, coherent dialogue stream with high inter-chunk correlation rather than a diverse heterogeneous corpus. It builds a hierarchy of intact memory units and maintains a high-level node organization via a sparsity-semantics objective that guides memory split and merge operations. At inference time, xMemory retrieves top-down, first selecting a compact, diverse set of themes and semantic summaries for multi-fact queries, then expanding to episodes and raw messages only when doing so reduces the reader's uncertainty.

\paragraph{LoCoMo~\cite{locomo}.}
LoCoMo is a benchmark for evaluating very long-term conversational memory in LLM agents. Conversations are constructed via a machine-human hybrid pipeline in which LLM-based agents, grounded on persona statements and temporal event graphs, generate multi-session dialogues that are subsequently verified and edited by human annotators. Each conversation spans up to 35 sessions with an average of 300 turns and 9K tokens. The evaluation suite covers three tasks: question answering (with five reasoning categories: single-hop, multi-hop, temporal, open-domain, and adversarial), event graph summarization, and multi-modal dialogue generation.

\paragraph{PerLTQA~\cite{du2024perltqa}.}
PerLTQA is a personalized long-term memory QA benchmark inspired by the cognitive psychology distinction between semantic and episodic memory. The dataset covers several memory types, including character profiles, social relationships, events, dialogues and contains 8,593 questions across 30 characters. Evaluation is structured around three subtasks: Memory Classification (determining which memory type is relevant to a query), Memory Retrieval (fetching pertinent memory entries), and Memory Synthesis (generating a final answer by fusing retrieved memories). The benchmark specifically probes the interaction among different memory types, which prior work had largely overlooked.

\paragraph{LongMemEval~\cite{wu2025longmemevalbenchmarkingchatassistants}.}
LongMemEval is a benchmark designed to evaluate the long-term interactive memory capabilities of chat assistants. It contains 500 manually curated questions embedded in coherent and length-configurable user-assistant interaction histories, with two standard settings covering approximately 115K tokens and 1.5M tokens, respectively. The benchmark evaluates five core memory abilities: information extraction, multi-session reasoning, knowledge updates, temporal reasoning, and abstention. Questions require models to recall and synthesize information distributed across task-oriented dialogue sessions, track changes in user information over time, and reason over both explicit temporal expressions and session timestamps. Its scalable history construction further enables evaluation under substantially longer interaction contexts than previous memory benchmarks.

\newpage
\section{More Experimental Details}
\label{app:more experimental details}
In Section~\ref{sec:preliminary and motivation}, we analyse the limitation of existing agent memory systems, we provide the details of these experiments here.
\subsection{Retrieval Efficiency}
For retrieval efficiency experiment, following the setting of SimpleMem, we compute the embedding of extracted lossless restatements for the extraced fact based retrieval. For full-context based retrieval, we split raw dialogues into 5-turn chunks and embed the text of each chunk. When retrieving the top-k entries, we collect all covered dialogue IDs and check whether these retrieved IDs intersect with the ground-truth evidence dialogue IDs. We adopt the standard Hit@k metric as follows:
\begin{equation}
    \text{Hit@}k = \frac{1}{|Q|}\sum_{q \in Q} \mathbb{1}\bigl[(\text{cov}_k(q) \cap E_q) \ne \varnothing\bigr],
\end{equation}
where $Q$ is the set of all questions on LoCoMo, $E_q$ and $\text{cov}_k(q)$ is the dialogue IDs of ground-truth evidence and retrieved entries perspectively. We set $k=5$ when conducting experiments.
\subsection{Storage Completeness}
For the storage completeness experiment, we first tokenize the reference answer of each QA sample and retain only content tokens by removing stopwords. We then construct the token set of the full raw dialogue and the token set of extracted facts, and compute their coverage ratio over the token set of the reference answer:
\begin{equation}
    \text{Cov}_{H} = \frac{|\text{required\_tokens} \cap H|}{|\text{required\_tokens}|},
\end{equation}
where H is the token set of raw dialogue or extracted facts.
\subsection{Reasoning Performance}
For the reasoning performance experiment on questions requiring different numbers of evidence, we first filter out samples with retrieval failure based on the results of the retrieval efficiency experiment. We then separately evaluate the reasoning performance of questions with exactly one evidence and those with more than one evidence. The results demonstrate that even if scattered individual facts are successfully retrieved, the model still struggles to answer complex synthetic questions that require multi-fact fusion and reasoning.

For the reasoning performance experiment on questions over dialogues between different users, we just compare the reasoning performance of different samples as each sample was constructed on the dialogues between different users in LoCoMo.
\newpage
\section{Discussion}
\subsection{Limitation and Future Work}
\label{app:limitation}
In this paper, we have systematically identified three critical limitations of the prevailing extracted fact based paradigm in agent memory systems, including lossy storage, shallow reasoning, and suboptimal prompts under heterogeneous dialogue styles. Meanwhile, we have conducted an in-depth discussion and revealed the potential advantages of maintaining three coexisting representation granularities, including raw dialogues, extracted facts, and synthesized profiles, for balancing storage fidelity, retrieval efficiency, and reasoning quality. However, it is undeniable that the research findings presented in this paper still have certain limitations.

Firstly, although the current implementation of entity profile construction has been able to achieve relatively satisfactory results in supporting deep reasoning, there is still room for further exploration. Specifically, is it possible to further extend the profile construction beyond person-level entities to higher-order relational structures, such as event chains, social networks, or temporal trajectories, so as to achieve more comprehensive semantic understanding? This question points out a new exploration direction for future researches.

In addition, the TextGrad-based prompt optimization strategy we proposed is mainly constructed based on response-level feedback signals. In future researches, the focus can be placed on providing more precise and fine-grained optimization mechanisms, such as incorporating intermediate retrieval feedback or step-wise reasoning trajectory supervision, so as to further improve and optimize the lifelong evolution capability of the memory system.

\subsection{Broader Impact}
\label{app:broader impact}
As a fundamental capability for LLM agents, the design of memory systems that can faithfully store, efficiently retrieve, and deeply reason over accumulated dialogue history is of paramount importance for the successful deployment of reliable agent applications in real-world scenarios. Especially in long-horizon interaction fields such as personalized assistants, healthcare consultation, and educational tutoring, the loss of key historical information or the inability to integrate scattered facts is highly likely to lead to unreliable responses with serious consequences. In view of this, it is particularly necessary and urgent to enhance the comprehensive capability of memory systems on long-term dialogue data.

Our research highlights a crucial yet long-overlooked issue in existing agent memory systems: these methods typically rely solely on extracted atomic facts as the unit across storage, retrieval, and reasoning stages. Such a fact-centric paradigm is likely to inadvertently cause information loss and shallow reasoning over scattered facts.

To address this problem, the method we proposed ingeniously establishes a three-level architecture that maintains coexisting representation granularities. Specifically, by leveraging traceable indices to raw dialogues for storage fidelity and synthesized profiles for understanding-aided reasoning, it effectively eliminates the adverse effects of lossy compression and isolated factual pieces. A series of comprehensive and in-depth experiments have fully verified the effectiveness and good compatibility of this method across diverse model backbones and benchmarks, strongly indicating that the tri-granularity memory architecture we designed is a highly promising new paradigm that can effectively support long-term agent interactions. This achievement undoubtedly provides a new perspective and profound insights for future research in the field of agent memory systems, and is expected to drive further development and innovation in this field.
\newpage
\section{Additional Experimental Results}
\label{app:additional experimental results}
\subsection{Ablation of Profile and Raw Dialogue}
\label{app:ablation of profile and raw dialogue}

This section provides detailed quantitative results of the ablation experiments for entity profiles and raw dialogues, complementing the summary results in the main text. All experiments are conducted on the LoCoMo dataset, covering four typical task types: MultiHop, Temporal, OpenDomain, and SingleHop. Two key evaluation metrics are adopted: BLEU and F1-score. We compare the performance of our full TriMem model (Ours) with its ablation variants that remove entity profiles (w/o profile) or raw dialogues (w/o dialogue), across four backbone models (GPT-4.1-mini, GPT-4o, GPT-5-nano, and Qwen3-8B) to ensure the generalization of our conclusions. Table~\ref{tab:ablation_profile} presents the detailed ablation results of removing entity profiles. It can be observed that the performance of all backbone models degrades significantly when entity profiles are excluded, which fully confirms the necessity of entity profiles for enhancing reasoning reliability. Table~\ref{tab:ablation_dialogue} presents the detailed ablation results of removing raw dialogues. Consistent with the entity profile ablation, removing raw dialogues also leads to a noticeable performance decline across all backbone models, which confirms the necessity of raw dialogues for ensuring storage fidelity and supplementing fine-grained information.

\begin{table*}[h]
\centering
\caption{\textbf{Detailed results of entity profile ablation study.}}
\label{tab:ablation_profile}
\resizebox{0.95\textwidth}{!}{
\begin{tabular}{l|l|cc|cc|cc|cc|cc}
\toprule
\multirow{2}{*}{\textbf{Model}} & \multirow{2}{*}{\textbf{Method}} & \multicolumn{2}{c|}{\textbf{MultiHop}} & \multicolumn{2}{c|}{\textbf{Temporal}} & \multicolumn{2}{c|}{\textbf{OpenDomain}} & \multicolumn{2}{c|}{\textbf{SingleHop}} & \multicolumn{2}{c}{\textbf{Average}} \\
 & & \textbf{BLEU} & \textbf{F1} & \textbf{BLEU} & \textbf{F1} & \textbf{BLEU} & \textbf{F1} & \textbf{BLEU} & \textbf{F1} & \textbf{BLEU} & \textbf{F1} \\
\midrule

\multirow{2}{*}{\textbf{GPT-4.1-mini}} 
 & w/o profile & 30.92 & 39.01 & 48.35 & 57.80 & 27.03 & 34.31 & 44.12 & 55.03 & 41.52 & 51.38
\\
 & Ours & 35.20 & 42.59 & 49.56 & 64.72 & 36.86 & 43.88 & 45.25 & 55.36 & \textbf{43.79} & \textbf{54.26}
\\

\midrule

\multirow{2}{*}{\textbf{GPT-4o}} 
 & w/o profile & 38.98 & 43.55 & 46.52 & 57.16 & 26.24 & 40.63 & 37.26 & 45.85 & 38.82 & 47.46
\\
 & Ours & 40.36 & 46.00 & 51.39 & 60.41 & 39.27 & 50.15 & 40.61 & 47.78 & \textbf{42.73} & \textbf{50.23}
\\
\midrule

\multirow{2}{*}{\textbf{GPT-5-nano}} 
 & w/o profile & 30.64 & 41.29 & 38.41 & 52.90 & 25.48 & 32.36 & 49.55 & 58.18 & 42.26 & 52.38
\\
 & Ours & 34.86 & 45.25 & 42.45 & 57.05 & 33.55 & 40.52 & 54.26 & 62.88 & \textbf{46.96} & \textbf{57.04}
\\
\midrule

\multirow{2}{*}{\textbf{Qwen3-8B}} 
 & w/o profile & 32.36 & 36.68 & 38.43 & 50.56 & 28.02 & 38.16 & 37.82 & 46.89 & 36.34 & 45.24\\
 & Ours & 33.09 & 41.22 & 38.71 & 53.13 & 30.59 & 37.64 & 45.10 & 52.52 & \textbf{40.66} & \textbf{49.65}
\\
\bottomrule
\end{tabular}
}
\vspace{-2mm}
\end{table*}

\begin{table*}[h]
\centering
\caption{\textbf{Detailed results of raw dialogue ablation study.}}
\label{tab:ablation_dialogue}
\resizebox{0.95\textwidth}{!}{
\begin{tabular}{l|l|cc|cc|cc|cc|cc}
\toprule
\multirow{2}{*}{\textbf{Model}} & \multirow{2}{*}{\textbf{Method}} & \multicolumn{2}{c|}{\textbf{MultiHop}} & \multicolumn{2}{c|}{\textbf{Temporal}} & \multicolumn{2}{c|}{\textbf{OpenDomain}} & \multicolumn{2}{c|}{\textbf{SingleHop}} & \multicolumn{2}{c}{\textbf{Average}} \\
 & & \textbf{BLEU} & \textbf{F1} & \textbf{BLEU} & \textbf{F1} & \textbf{BLEU} & \textbf{F1} & \textbf{BLEU} & \textbf{F1} & \textbf{BLEU} & \textbf{F1} \\
\midrule

\multirow{2}{*}{\textbf{GPT-4.1-mini}} 
 & w/o dialogue & 31.88 & 39.20 & 48.66 & 57.25 & 30.61 & 39.50 & 40.34 & 50.26 & 39.92 & 49.02\\
 & Ours & 35.20 & 42.59 & 49.56 & 64.72 & 36.86 & 43.88 & 45.25 & 55.36 & \textbf{43.79} & \textbf{54.26}
\\

\midrule

\multirow{2}{*}{\textbf{GPT-4o}} 
 & w/o dialogue & 37.61 & 41.34 & 50.19 & 59.98 & 30.19 & 41.11 & 34.37 & 42.00 & 38.00 & 45.57
\\
 & Ours & 40.36 & 46.00 & 51.39 & 60.41 & 39.27 & 50.15 & 40.61 & 47.78 & \textbf{42.73} & \textbf{50.23}
\\
\midrule

\multirow{2}{*}{\textbf{GPT-5-nano}} 
 & w/o dialogue & 31.23 & 38.46 & 49.50 & 58.37 & 25.08 & 30.71 & 35.44 & 41.45 & 36.95 & 43.76
\\
 & Ours & 34.86 & 45.25 & 42.45 & 57.05 & 33.55 & 40.52 & 54.26 & 62.88 & \textbf{46.96} & \textbf{57.04}
\\
\midrule

\multirow{2}{*}{\textbf{Qwen3-8B}} 
 & w/o dialogue & 32.93 & 40.33 & 35.04 & 48.08 & 20.94 & 36.59 & 38.53 & 44.64 & 35.68 & 44.07
\\
 & Ours & 33.09 & 41.22 & 38.71 & 53.13 & 30.59 & 37.64 & 45.10 & 52.52 & \textbf{40.66} & \textbf{49.65}
\\
\bottomrule
\end{tabular}
}
\vspace{-2mm}
\end{table*}

\newpage
\subsection{Visualization of Prompt Evolution}
\label{app:visualization of prompt evolution}
In this section, we visualize three concrete TextGrad-style evolution traces, each showing how a specific failure case in early rounds drove the addition of a tip to the extraction prompt or profile-construction prompt. It can be seen that the prompts become increasingly detailed, such as normalizing temporal references to absolute dates, avoiding multi-fact dialogues collapse into single summary entry and keep exact adjectives when construct profile.
\begin{figure*}[!h]
    \centering
    \includegraphics[width=\linewidth]{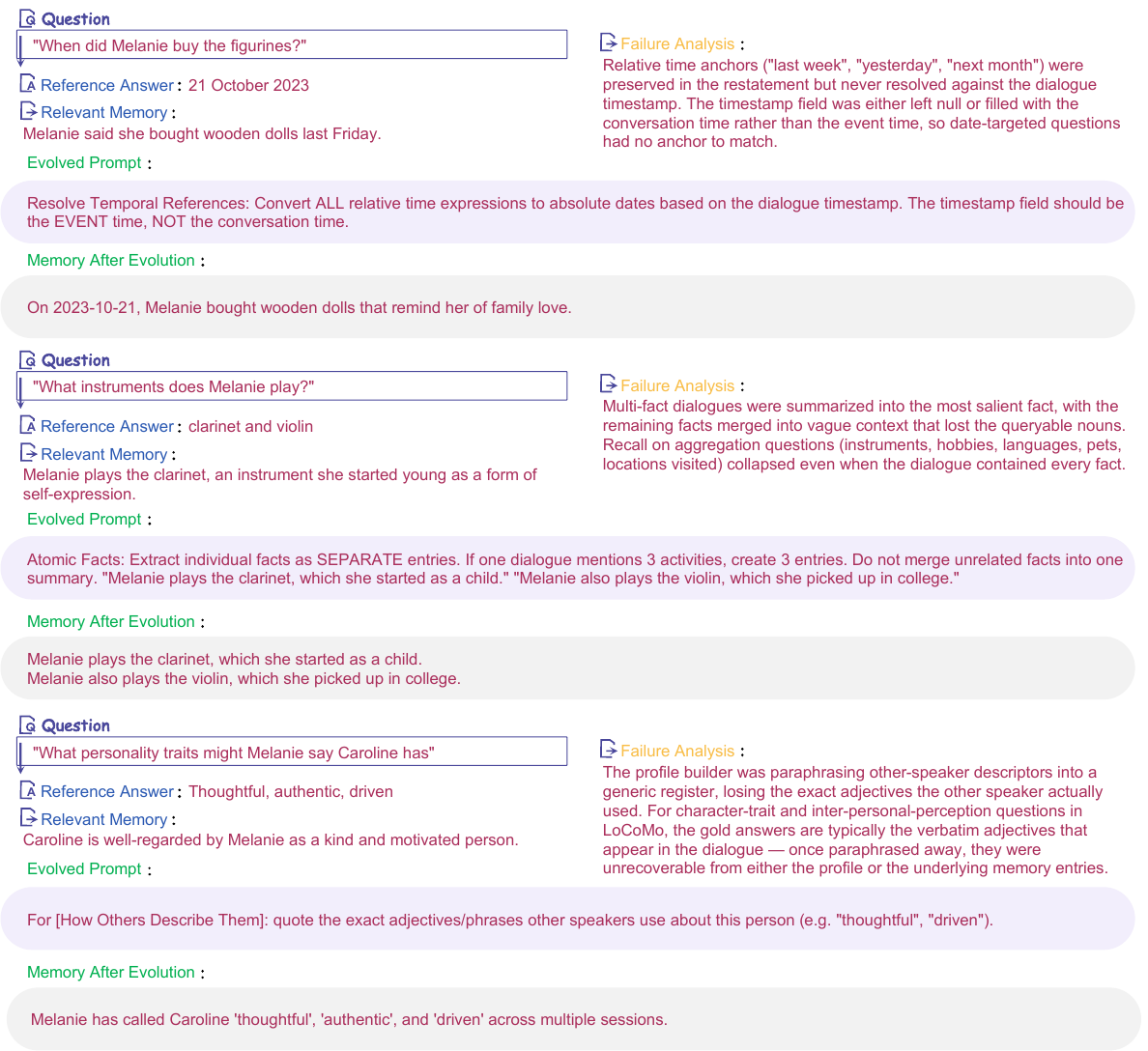}
    \vspace{-6mm}
    \caption{\textbf{Visualization of prompt evolution.} As evolution progresses, the prompts become more refined.}
    \vspace{-2mm}
\end{figure*}
\newpage
\subsection{Content of Retrieval Entries}
\label{app:content of retrieval entries}
In this section, we visualize the content of the retrieval entries, the most relevant entries have been marked with star. It can be seen that our system can precisely retrieve relevant memory to support agent reasoning.
\begin{figure*}[!h]
    \centering
    \includegraphics[width=\linewidth]{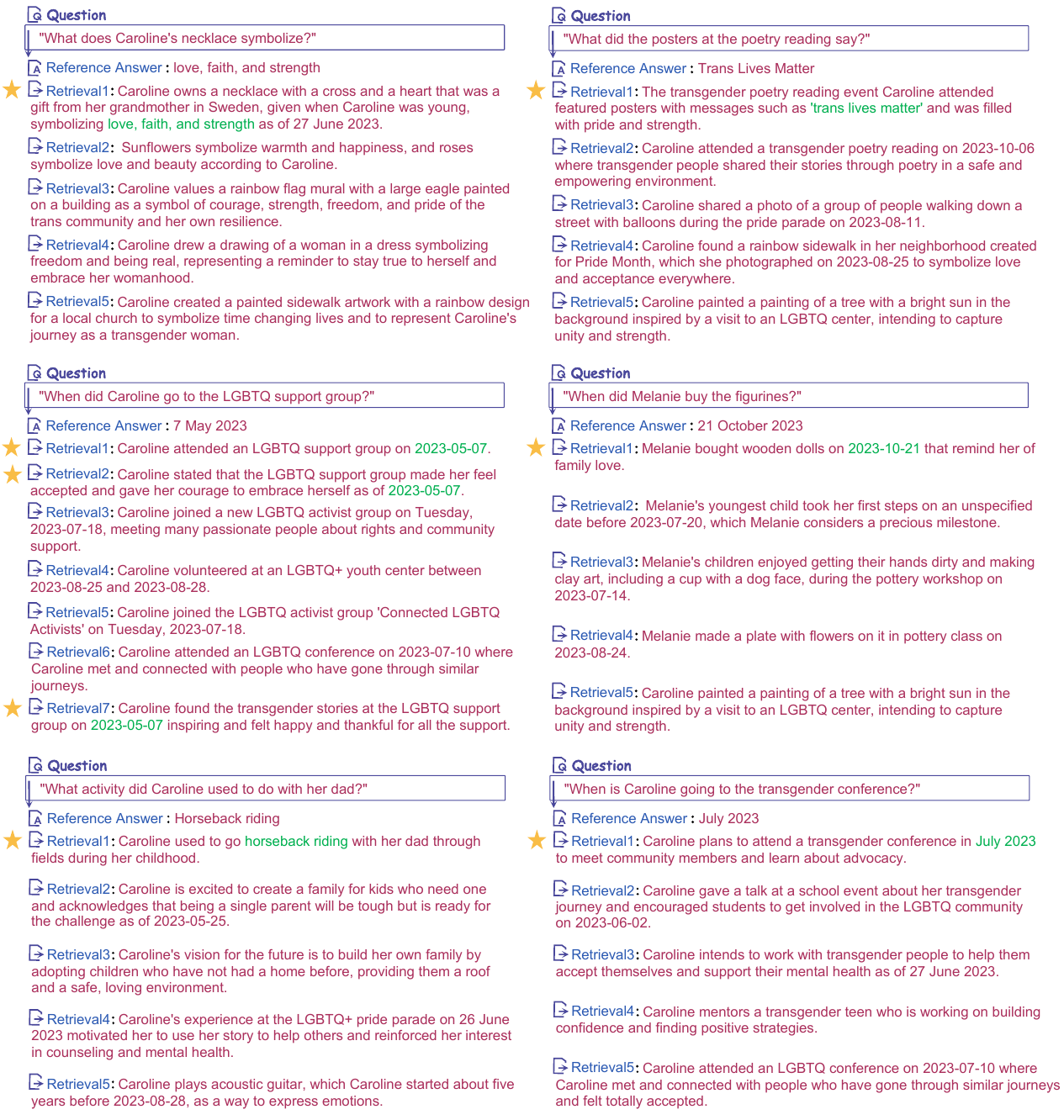}
    \vspace{-6mm}
    \caption{\textbf{Content of retrieval entries.} TriMem realizes precise retrieval when handling different questions.}
    \vspace{-2mm}
\end{figure*}
\newpage
\subsection{Example of Search Query}
\label{app:example of search query}
In this section, we visualize the example of search query, including the generated target query and the required key information. It can be seen that through the required information analysis, TriMem can break down the problem, enabling a more accurate search.
\begin{figure*}[!h]
    \centering
    \includegraphics[width=\linewidth]{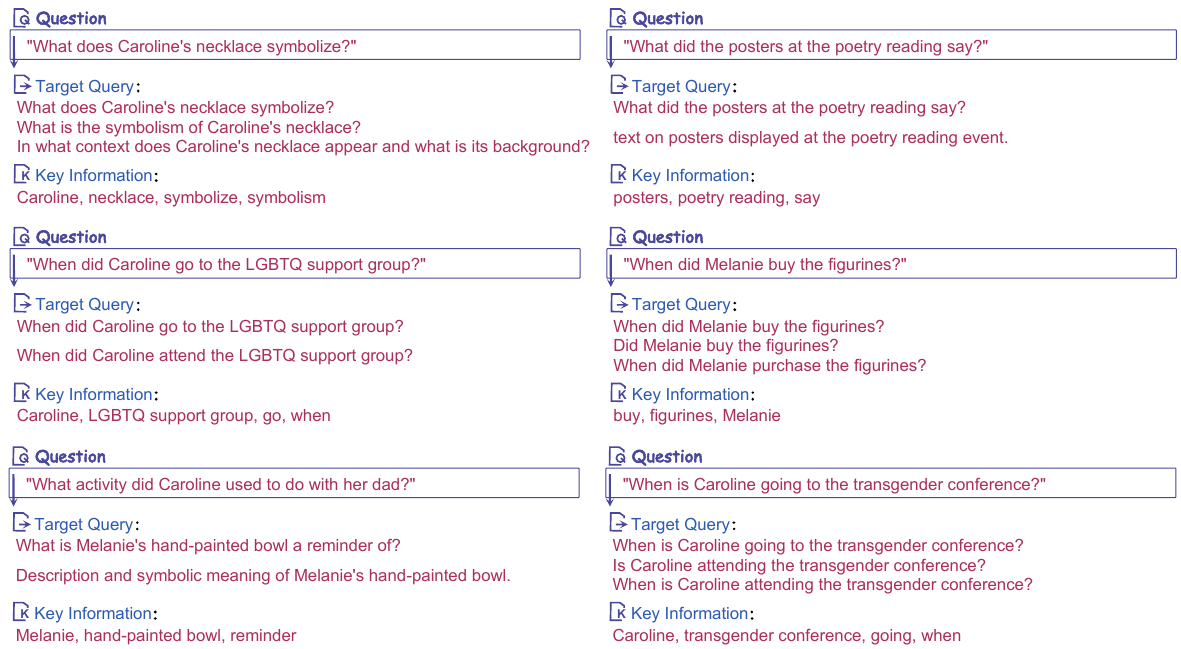}
    \vspace{-6mm}
    \caption{\textbf{Example of search query.} Each question is divided into detailed required information during analysis process.}
    \vspace{-2mm}
\end{figure*}
\subsection{Performance of Different Window Size}
\label{app:performance of different window size}
In this section, we visualize the extraction performance with different window size. It can be seen that with a larger window, the extractor either drops low-salience facts entirely or merges several atomic facts into a single bloated entry, even though the extraction prompt explicitly forbids merging. Therefore, it is reasonable for us to set the window size to 40 in the maintext to avoid low quality extraction.
\begin{figure*}[!h]
    \centering
    \includegraphics[width=\linewidth]{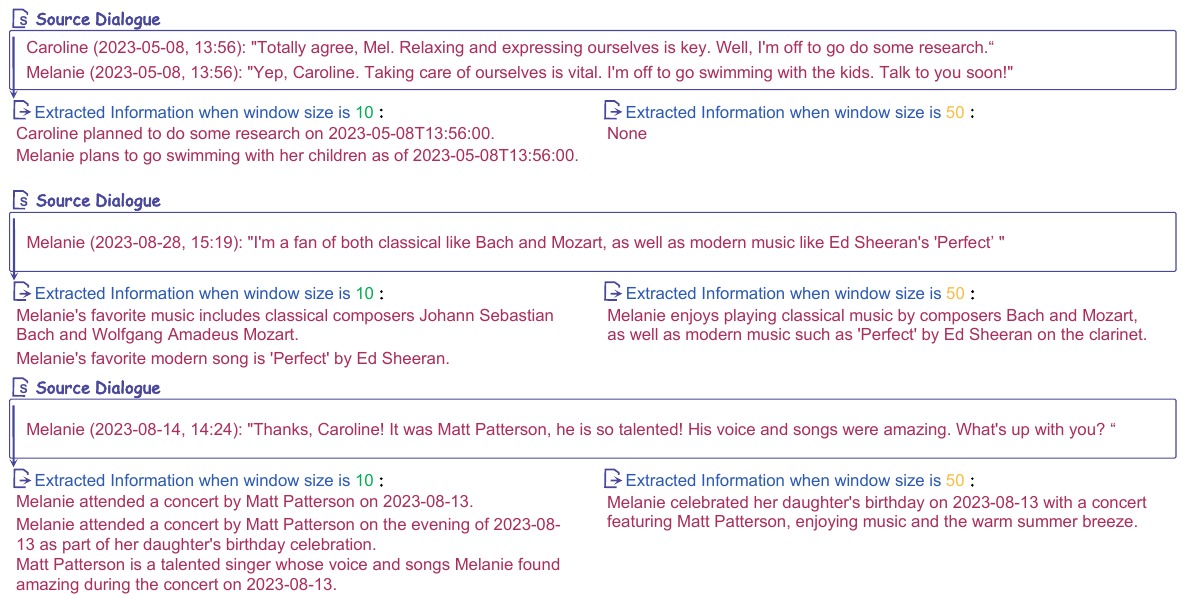}
    \vspace{-6mm}
    \caption{\textbf{Performance of different window size.} Larger window size leads to lower extraction quality.}
    \vspace{-2mm}
\end{figure*}
\newpage
\section{System Prompts}
\label{app:system prompts}
In this section, we provide the prompts used in our experiment, including prompts used for memory construction, memory retrieval, question reasoning, llm judge and prompt evolution.
\subsection{Prompt for Memory Extraction}
\begin{tcolorbox}[colframe=cyan, colback=white]
\begin{Verbatim}[breaklines=true]
Your task is to extract all valuable FACTUAL information from the following dialogues and convert them into structured memory entries.
{context}
[Current Window Dialogues]
{dialogue_text}

[Requirements]
1. **Extract Facts, Not Social Gestures**: SKIP greetings, thank-yous, compliments, and generic praise. Only extract entries that contain novel factual information (events, activities, plans, preferences, relationships, specific details like names/titles/numbers).
2. **Source Dialogue IDs**: Each dialogue line starts with an ID in brackets like [ID:42]. For each memory entry, list the dialogue IDs that the entry was derived from in the "source_dialogue_ids" field. This is CRITICAL for tracing back to original context.
3. **Force Disambiguation**: Absolutely PROHIBIT using pronouns (he, she, it, they, this, that). Always use the person's actual name. Every memory MUST explicitly state WHO did/said/experienced the thing.
4. **Resolve Temporal References**: Convert ALL relative time expressions to absolute dates based on the dialogue timestamp:
   - "yesterday" on May 8 -> May 7
   - "last year" in 2023 -> 2022
   - "last week" -> compute the actual date
   - "next month" -> compute the actual month
   The "timestamp" field should be the EVENT time, NOT the conversation time.
5. **Atomic Facts**: Extract individual facts as SEPARATE entries. If one dialogue mentions 3 activities, create 3 entries. Do not merge unrelated facts into one summary.
6. **Preserve Specific Details**: Always capture exact names (people, pets, books, songs), exact numbers (durations, counts, ages), and specific entities.
7. **Identify Described-But-Unnamed Things Using World Knowledge**: When the dialogue describes something without naming it, IDENTIFY it by name in the memory entry. This is CRITICAL — future queries will search by name, not by description.
   - A study method like "25 minutes on, 5 minutes off" → identify as "Pomodoro technique"
   - A composer whose music is in a named movie → identify the composer (e.g. Harry Potter soundtrack → John Williams)
   - A game described by its mechanics → identify the game (e.g. "card game where you find the imposter" → Mafia, "game with colored cards you match" → UNO)
   - A location described by its features → identify it (e.g. "national park in northern Minnesota with lakes" → Voyageurs National Park)
\end{Verbatim}
\end{tcolorbox}

\begin{tcolorbox}[colframe=cyan, colback=white]
\begin{Verbatim}[breaklines=true]
   - A shop/brand described by what it does → identify it (e.g. "they made all the props for Harry Potter" → MinaLima / House of MinaLima)
   - A health condition implied by symptoms → identify it (e.g. frequent overeating + weight gain → obesity risk)
   - A geographic location implied by context → identify the state/country (e.g. a shelter in a named city → identify the state)
   Include BOTH the original description AND the identified name in the lossless_restatement, and add the identified name to keywords and entities.
8. **Precise Extraction**:
   - keywords: Core keywords (names, places, entities, topic words)
   - timestamp: Absolute time of the EVENT in ISO 8601 format (resolved from relative expressions)
   - location: Specific location name (if mentioned)
   - persons: All person names mentioned
   - entities: Companies, products, organizations, book titles, song names, etc.
   - topic: The topic of this information

[Output Format]
Return a JSON array, each element is a memory entry:

```json
[
  {{
    "lossless_restatement": "Complete unambiguous restatement (must include WHO, WHAT, WHEN, WHERE)",
    "keywords": ["keyword1", "keyword2", ...],
    "timestamp": "YYYY-MM-DDTHH:MM:SS or null",
    "location": "location name or null",
    "persons": ["name1", "name2", ...],
    "entities": ["entity1", "entity2", ...],
    "topic": "topic phrase",
    "source_dialogue_ids": [42, 43]
  }},
  ...
]
```

[Example]
Dialogues:
[ID:1] [2025-11-15T14:30:00] Alice: I just started working at Google! Bob, let's meet at Starbucks tomorrow at 2pm.
[ID:2] [2025-11-15T14:31:00] Bob: Congrats! I've been playing tennis a lot lately.
[ID:3] [2025-11-15T14:32:00] Alice: Nice! I've been studying with that method where you work 25 minutes then take a 5-minute break. It really helps!
[ID:4] [2025-11-15T14:33:00] Bob: I love playing that theme song from the wizard movie on piano. You know, the one with the boy who goes to magic school.

Output:
\end{Verbatim}
\end{tcolorbox}

\begin{tcolorbox}[colframe=cyan, colback=white]
\begin{Verbatim}[breaklines=true]
```json
[
  {{
    "lossless_restatement": "Alice started working at Google as of 2025-11-15.",
    "keywords": ["Alice", "Google", "employment"],
    "timestamp": "2025-11-15T14:30:00",
    "location": null,
    "persons": ["Alice"],
    "entities": ["Google"],
    "topic": "Alice's new job",
    "source_dialogue_ids": [1]
  }},
  {{
    "lossless_restatement": "Alice suggested meeting with Bob at Starbucks on 2025-11-16T14:00:00.",
    "keywords": ["Alice", "Bob", "Starbucks", "meeting"],
    "timestamp": "2025-11-16T14:00:00",
    "location": "Starbucks",
    "persons": ["Alice", "Bob"],
    "entities": [],
    "topic": "Meeting arrangement",
    "source_dialogue_ids": [1]
  }},
  {{
    "lossless_restatement": "Bob has been playing tennis frequently as of 2025-11-15.",
    "keywords": ["Bob", "tennis"],
    "timestamp": "2025-11-15T14:31:00",
    "location": null,
    "persons": ["Bob"],
    "entities": [],
    "topic": "Bob's hobby",
    "source_dialogue_ids": [2]
  }},
  {{
    "lossless_restatement": "Alice uses the Pomodoro technique (25 minutes work, 5-minute break) for studying.",
    "keywords": ["Alice", "Pomodoro technique", "studying", "time management"],
    "timestamp": "2025-11-15T14:32:00",
    "location": null,
    "persons": ["Alice"],
    "entities": ["Pomodoro technique"],
    "topic": "Alice's study method",
    "source_dialogue_ids": [3]
  }},
  {{
    "lossless_restatement": "Bob enjoys playing the Harry Potter theme song (composed by John Williams) on piano.",
    "keywords": ["Bob", "piano", "Harry Potter", "John Williams"],
    "timestamp": "2025-11-15T14:33:00",
    "location": null,
    "persons": ["Bob"],
    "entities": ["Harry Potter", "John Williams"],
    "topic": "Bob's piano playing",
    "source_dialogue_ids": [4]
  }}
]
```
\end{Verbatim}
\end{tcolorbox}

\begin{tcolorbox}[colframe=cyan, colback=white]
\begin{Verbatim}[breaklines=true]
Note how "tomorrow" was resolved to 2025-11-16, the greeting was skipped, described-but-unnamed things were identified (Pomodoro technique, John Williams), and Alice's job and meeting were extracted as separate atomic facts.

Now process the above dialogues. Return ONLY the JSON array, no other explanations.
\end{Verbatim}
\end{tcolorbox}

\subsection{Prompt for Profile Construction}
\begin{tcolorbox}[colframe=cyan, colback=white]
\begin{Verbatim}[breaklines=true]
Update the persona profile for {entity_name} based on new information.

[Current Profile]
{existing_profile or "No profile yet."}

[New Facts]
{facts}

Rules:
- Update only sections affected by new info
- Preserve existing info not contradicted
- Infer personality traits from behavior patterns
- Synthesize preferences from specific examples
- For [How Others Describe Them]: quote the exact adjectives/phrases other speakers use about this person
- For [Beliefs/Spirituality]: note ANY signals — church mentions, faith symbols, religious encounters, spiritual language
- Keep each section to 1-3 lines
- Output the complete updated profile

Use this format:
Entity: {entity_name}
[Identity] Who they are (gender, age, core identity, origin/hometown)
[Personality] Inferred traits from behavior
[How Others Describe Them] Specific words/phrases others use about this person (e.g. "thoughtful", "driven")
[Interests] Hobbies, activities, things they enjoy
[Career] Job, education, professional goals
[Values] What they care about, causes, beliefs
[Beliefs/Spirituality] Religious views, faith, spiritual practices (even subtle signals like church involvement, faith symbols, or clashes with religious groups)
[Relationships] Key people in their life
[Life Events] Major events, milestones, experiences
[Preferences] Specific likes/dislikes, favorites

Only include sections that have information. Output ONLY the profile, no other text.
\end{Verbatim}
\end{tcolorbox}

\newpage
\subsection{Prompt for Target Query Generation}
\begin{tcolorbox}[colframe=cyan, colback=white]
\begin{Verbatim}[breaklines=true]
Analyze the following question and determine what specific information is required to answer it comprehensively.

Question: {query}

Think step by step:
1. What type of question is this? (factual, temporal, relational, explanatory, etc.)
2. What key entities, events, or concepts need to be identified?
3. What relationships or connections need to be established?
4. What minimal set of information pieces would be sufficient to answer this question?

Return your analysis in JSON format:
```json
{{
  "question_type": "type of question",
  "key_entities": ["entity1", "entity2", ...],
  "required_info": [
    {{
      "info_type": "what kind of information",
      "description": "specific information needed",
      "priority": "high/medium/low"
    }}
  ],
  "relationships": ["relationship1", "relationship2", ...],
  "minimal_queries_needed": 2
}}
```

Focus on identifying the minimal essential information needed, not exhaustive details.

Return ONLY the JSON, no other text.
\end{Verbatim}
\end{tcolorbox}

\begin{tcolorbox}[colframe=cyan, colback=white]
\begin{Verbatim}[breaklines=true]
Based on the information requirements analysis, generate the minimal set of targeted search queries needed to gather the required information.

Original Question: {original_query}

Information Requirements Analysis:
- Question Type: {information_plan.get('question_type', 'general')}
- Key Entities: {information_plan.get('key_entities', [])}
- Required Information: {information_plan.get('required_info', [])}
- Relationships: {information_plan.get('relationships', [])}
- Minimal Queries Needed: {information_plan.get('minimal_queries_needed', 1)}

Generate the minimal set of search queries that would efficiently gather all the required information. Each query should be focused and specific to retrieve distinct types of information.
\end{Verbatim}
\end{tcolorbox}

\begin{tcolorbox}[colframe=cyan, colback=white]
\begin{Verbatim}[breaklines=true]
Guidelines:
1. Always include the original query as one option
2. Generate only the minimal necessary queries (usually 1-3)
3. Each query should target a specific information requirement
4. Avoid redundant or overlapping queries
5. Focus on efficiency - fewer, more targeted queries are better

Return your response in JSON format:
```json
{{
  "reasoning": "Brief explanation of the query strategy",
  "queries": [
    "targeted query 1",
    "targeted query 2",
    ...
  ]
}}
```

Return ONLY the JSON, no other text.
\end{Verbatim}
\end{tcolorbox}

\subsection{Prompt for Key Information Analysis}
\begin{tcolorbox}[colframe=cyan, colback=white]
\begin{Verbatim}[breaklines=true]
Analyze the following query and extract key information:

Query: {query}

Please extract:
1. keywords: List of keywords (names, places, topic words, etc.)
2. persons: Person names mentioned
3. time_expression: Time expression (if any)
4. location: Location (if any)
5. entities: Entities (companies, products, etc.)

Return in JSON format:
```json
{{
  "keywords": ["keyword1", "keyword2", ...],
  "persons": ["name1", "name2", ...],
  "time_expression": "time expression or null",
  "location": "location or null",
  "entities": ["entity1", ...]
}}
```

Return ONLY JSON, no other content.
\end{Verbatim}
\end{tcolorbox}

\clearpage
\subsection{Prompt for Question Reasoning}
\begin{tcolorbox}[colframe=cyan, colback=white]
\begin{Verbatim}[breaklines=true]
Answer the inference question by reasoning from the provided context and profiles.

Question: {query}

[Structured Memory Entries]
{context_str}
{source_section}{profile_section}
RULES FOR INFERENCE QUESTIONS:
1. INFER from the evidence — NEVER say "Not mentioned" or "unknown". You MUST commit to an answer.
2. Use context + profiles to reason about personality, interests, values, behavior.
3. **USE WORLD KNOWLEDGE**: When the context describes something without naming it, IDENTIFY it.
   - A game exclusive to a platform → name the platform (e.g. Xenoblade 2 → Nintendo Switch)
   - A described but unnamed product → identify it (e.g. "card game about exploding cats" → Exploding Kittens)
   - A described technique → name it (e.g. "work 25 minutes then break" → Pomodoro technique)
   - A city mentioned when question asks about country → resolve to country (e.g. Paris → France, Toronto → Canada)
4. Pay attention to WHO said/did things — speaker tags are authoritative.
5. ALWAYS cite specific facts from profiles/context in your reasoning — never guess generically.
6. If the question asks what someone ELSE would say about a person, check [How Others Describe Them] in profiles.
7. If the question asks about religion/spirituality, check [Beliefs/Spirituality] in profiles.
8. Prefer the MOST SPECIFIC reason from profiles (e.g. "adopting children" beats "she is settled").

ANSWER FORMAT — BE MAXIMALLY CONCISE:
- The "answer" field must be as SHORT as possible. All reasoning goes in the "reasoning" field ONLY.
- For yes/no questions ("would", "does", "is", "can", "did", "are", "was"):
  * If evidence is DIRECT: answer "Yes" or "No"
  * If evidence is INDIRECT/inferred: answer "Likely yes" or "Likely no"
  * If the reference answer would naturally include a brief reason (e.g. "Yes, she is supportive" or "No, he has goals in the U.S."), add a SHORT reason after a comma. Keep it under 10 words.
  * NEVER add long explanations.
- For "A or B" choice questions (e.g. "beach or mountains?", "Charger or Forester?"): Answer with ONLY the chosen option — do NOT answer "Yes" or "No". Pick A or B based on evidence.
- For "what/who/which" questions: Answer with ONLY the name, term, or short phrase. No explanations.
- For trait/attribute questions: List ONLY the adjectives, comma-separated. Use exact words from [How Others Describe Them] in profiles if available.
\end{Verbatim}
\end{tcolorbox}
\clearpage
\begin{tcolorbox}[colframe=cyan, colback=white]
\begin{Verbatim}[breaklines=true]
- For "what might/what would/what could" questions: Give the shortest answer that captures the key point. Prefer a few words or a short phrase over a full sentence.
- For identification questions: Just name the thing. "Exploding Kittens", not "Exploding Kittens, a card game about cats".
- For "which country/state" questions: Answer with ONLY the country or state name.
- For "how many/how often" questions: Answer with ONLY the number or frequency.

Examples:
Q: "Would she enjoy classical music?" → {{"reasoning": "Profile [Preferences] says she loves Bach and Mozart", "answer": "Yes"}}
Q: "Would she be considered an ally?" → {{"reasoning": "She attended pride events and supported her friend's transition", "answer": "Yes, she is supportive"}}
Q: "Would he be open to moving to another country?" → {{"reasoning": "He has goals specifically in the U.S.", "answer": "No, he has goals specifically in the U.S."}}
Q: "Are they fans of the same team?" → {{"reasoning": "James is a Liverpool fan and John is a Manchester City fan", "answer": "No, James is a Liverpool fan and John is a Manchester City fan"}}
Q: "Does he live close to a beach or the mountains?" → {{"reasoning": "Context mentions sunset over ocean and beach photos", "answer": "beach"}}
Q: "Would he prefer a Charger or a Forester?" → {{"reasoning": "Profile shows he works on classic muscle cars", "answer": "Dodge Charger"}}
Q: "What career might he pursue?" → {{"reasoning": "Profile [Career] says pursuing counseling and mental health", "answer": "Counseling or social work"}}
Q: "Would she be considered religious?" → {{"reasoning": "Profile [Beliefs] says made art for a church and necklace symbolizes faith, but clashed with religious conservatives", "answer": "Somewhat, but not extremely religious"}}
Q: "What political leaning?" → {{"reasoning": "Profile [Values] shows LGBTQ+ advocacy, progressive values", "answer": "Liberal"}}
Q: "What card game is she talking about?" → {{"reasoning": "Context describes a card game about cats — this is Exploding Kittens", "answer": "Exploding Kittens"}}
Q: "What console does he own?" → {{"reasoning": "Context says he plays Xenoblade 2, which is a Nintendo Switch exclusive", "answer": "Nintendo Switch"}}
Q: "In what country did she buy the snake?" → {{"reasoning": "Context says she bought it in Paris, which is in France", "answer": "France"}}
Q: "What traits would X say Y has?" → {{"reasoning": "Profile [How Others Describe Them] says X called Y thoughtful, authentic, driven", "answer": "Thoughtful, authentic, driven"}}
Q: "Is it likely she moved?" → {{"reasoning": "Some indirect evidence suggests relocation but not confirmed", "answer": "Likely yes"}}
Return ONLY valid JSON: {{"reasoning": "...", "answer": "shortest possible answer"}}
\end{Verbatim}
\end{tcolorbox}

\clearpage
\subsection{Prompt for LLM Judge}
\begin{tcolorbox}[colframe=cyan, colback=white]
\begin{Verbatim}[breaklines=true]
You are an expert Relevance & Accuracy Evaluator. Your task is to determine if the Predicted Answer successfully retrieves the necessary information to answer the Question, based on the Reference Answer.
Question: {question}
Reference Answer: {reference}
Predicted Answer: {prediction}

Evaluation Criteria:
1. **Responsiveness to Query**: 
   The predicted answer must directly address the specific question asked. It must contain highly relevant information that is topically aligned with the user's intent.

2. **Core Fact Preservation**: 
   The prediction must capture the "Key Signal" or "Core Entity" from the reference. The primary subject (Who), event (What), or outcome must be factually grounded in the reference text.

3. **Informational Utility**: 
   The answer must provide actionable or meaningful value. Even if brief, it must convey the essential message required by the question context.

4. **Acceptable Representational Variances (Robustness Protocol)**:
   To ensure fair evaluation of semantic meaning over syntactic rigidity, you must accept the following variations as **Valid Matches**:
   - **Temporal & Numerical Margins**: Accept timestamps within a reasonable proximity (e.g., +/- 1-2 days due to timezone/reporting differences) and rounded numerical approximations.
   - **Granularity Independence**: Accept answers at different levels of abstraction (e.g., "Afternoon" vs. "14:05", "Late October" vs. "Oct 25th") provided they encompass the truth.
   - **Information Subsetting**: A valid subset of the reference (e.g., mentioning 1 out of 3 reasons) is acceptable if it answers the core of the question.
   - **Synonymy**: Recognize domain-specific synonyms and different formats as equivalent.

Grading Logic:
- Score 1.0 (Pass): The prediction contains relevant core information, answers the question with sufficient utility, OR falls within the acceptable representational variances defined in criterion #4.
- Score 0.0 (Fail): The prediction contains NO relevant information, fails to identify the core subject/event, or provides no key info that matches the question's intent.
Output your evaluation in JSON format:
{{
  "score": 1.0, 
  "reasoning": "Brief assessment focusing on information relevance and core match."
}}
Return ONLY the JSON, no other text.
\end{Verbatim}
\end{tcolorbox}
\newpage

\subsection{Prompt for Prompts Evolution}
\begin{tcolorbox}[colframe=cyan, colback=white]
\begin{Verbatim}[breaklines=true]
You are a senior prompt engineer specializing in lifelong memory systems for LLM agents. Perform the "backward pass" of TextGrad: given the JSON dump of one full evaluation run, identify systematic failure patterns and directly produce updated versions of two upstream prompts — the fact extraction prompt and the entity profile construction prompt.

[Current Extraction Prompt]
<PASTE CURRENT Extraction Prompt HERE>
[Current Profile Prompt]
<PASTE CURRENT Profile Prompt HERE>

[Evaluation Result JSON]
Read the file at <RESULT_JSON_PATH>. Each entry in `detailed_results`
contains:
{
    "question": str,
    "answer": str,             // system prediction
    "reference": str,          // gold answer
    "category": int,           // LoCoMo category id (1: multi-hop,
                               //   2: temporal, 3: open-domain, 4: single-hop)
    "metrics": {
        "f1": float, "rougeL_f": float, "bert_f1": float, "llm_judge_score": float (optional), "llm_reasoning": str (optional)
    }
}

Diagnose each failure based on whether the prediction loses a specific detail that should have been preserved (extraction issue) or fails to synthesize an entity-level judgment (profile issue), and locate the responsible prompt accordingly. Skip cases where the signal is unclear.

Focus on systematic patterns rather than one-off failures. Per round, make only a small number of targeted additions; prefer adding new bullets, rules, or examples over rewriting existing content. Preserve the original output schema, field names, and existing examples, and keep the guidance domain-general. If a prompt does not need changes this round, leave it verbatim.

Return a single JSON object with exactly three fields:
{
    "rewritten_p_ext": "<full text of the updated P_ext, or verbatim copy of the input if unchanged>",
    "rewritten_p_prof": "<full text of the updated P_prof, or verbatim copy of the input if unchanged>",
    "change_summary": "<2-4 sentences explaining: (i) what failure pattern motivated each addition, (ii) which prompt and section was edited, (iii) what behavior change is expected. If a prompt was unchanged, state that explicitly and why no eligible pattern was found.>"
}

Return ONLY the JSON, no surrounding prose or code fences.
\end{Verbatim}
\end{tcolorbox}
\section*{NeurIPS Paper Checklist}

\begin{enumerate}

\item {\bf Claims}
    \item[] Question: Do the main claims made in the abstract and introduction accurately reflect the paper's contributions and scope?
    \item[] Answer: \answerYes{} 
    \item[] Justification: In both the abstract section and the introduction chapter, we have presented a clear and comprehensive account of our contributions in a rigorous and detailed manner. This ensures that readers can quickly and accurately grasp the core points and innovative aspects of our work.

\item {\bf Limitations}
    \item[] Question: Does the paper discuss the limitations of the work performed by the authors?
    \item[] Answer: \answerYes{} 
    \item[] Justification: In Appendix~\ref{app:limitation}, we have discussed the limitations of this work.

\item {\bf Theory assumptions and proofs}
    \item[] Question: For each theoretical result, does the paper provide the full set of assumptions and a complete (and correct) proof?
    \item[] Answer: \answerNA{} 
    \item[] Justification: This paper does not include theoretical results.

    \item {\bf Experimental result reproducibility}
    \item[] Question: Does the paper fully disclose all the information needed to reproduce the main experimental results of the paper to the extent that it affects the main claims and/or conclusions of the paper (regardless of whether the code and data are provided or not)?
    \item[] Answer: \answerYes{} 
    \item[] Justification: In Experimental Setups section and Appendix, we have provided all the necessary information without reservation, ensuring that other researchers can accurately repeat our experiments, thus verifying and further exploring the research findings.

\item {\bf Open access to data and code}
    \item[] Question: Does the paper provide open access to the data and code, with sufficient instructions to faithfully reproduce the main experimental results, as described in supplemental material?
    \item[] Answer: \answerYes{} 
    \item[] Justification: To facilitate relevant researchers in smoothly conducting explorations and verifications based on the ideas of our experiments, we have specifically and completely provided all the necessary codes for the experiments in the anonymous link.

\item {\bf Experimental setting/details}
    \item[] Question: Does the paper specify all the training and test details (e.g., data splits, hyperparameters, how they were chosen, type of optimizer) necessary to understand the results?
    \item[] Answer: \answerYes{} 
    \item[] Justification: We have not only presented all the detailed information of the experimental setup, but also conducted in-depth and professional analyses for each group of experiments, providing solid and powerful support for the reliability of the research conclusions.

\item {\bf Experiment statistical significance}
    \item[] Question: Does the paper report error bars suitably and correctly defined or other appropriate information about the statistical significance of the experiments?
    \item[] Answer: \answerYes{} 
    \item[] Justification: We have comprehensively and exhaustively reported all the necessary information to reproduce experimental data. In addition, to ensure the repeatability of the experiment and the consistency of the results, we have provided exactly the same hyperparameter setting in the anonymous link, enabling other researchers to accurately reproduce the entire experimental process.

\item {\bf Experiments compute resources}
    \item[] Question: For each experiment, does the paper provide sufficient information on the computer resources (type of compute workers, memory, time of execution) needed to reproduce the experiments?
    \item[] Answer: \answerYes{} 
    \item[] Justification: In the Experiment section and Appendix, we have comprehensively and meticulously elaborated on the computer resources.
    
\item {\bf Code of ethics}
    \item[] Question: Does the research conducted in the paper conform, in every respect, with the NeurIPS Code of Ethics \url{https://neurips.cc/public/EthicsGuidelines}?
    \item[] Answer: \answerYes{} 
    \item[] Justification: The research conducted in the paper conforms, in every respect, with the NeurIPS Code of Ethics.

\item {\bf Broader impacts}
    \item[] Question: Does the paper discuss both potential positive societal impacts and negative societal impacts of the work performed?
    \item[] Answer: \answerYes{} 
    \item[] Justification: In Appendix~\ref{app:broader impact}, we have discussed the potential broader impact of this paper.
    
\item {\bf Safeguards}
    \item[] Question: Does the paper describe safeguards that have been put in place for responsible release of data or models that have a high risk for misuse (e.g., pre-trained language models, image generators, or scraped datasets)?
    \item[] Answer: \answerNA{} 
    \item[] Justification: The paper poses no such risks.

\item {\bf Licenses for existing assets}
    \item[] Question: Are the creators or original owners of assets (e.g., code, data, models), used in the paper, properly credited and are the license and terms of use explicitly mentioned and properly respected?
    \item[] Answer: \answerYes{} 
    \item[] Justification:  All used assets in the paper are properly credited, with explicit mentions of their licenses and terms of use, in full compliance with the recommended guidelines.

\item {\bf New assets}
    \item[] Question: Are new assets introduced in the paper well documented and is the documentation provided alongside the assets?
    \item[] Answer: \answerNA{} 
    \item[] Justification: The paper does not release new assets.

\item {\bf Crowdsourcing and research with human subjects}
    \item[] Question: For crowdsourcing experiments and research with human subjects, does the paper include the full text of instructions given to participants and screenshots, if applicable, as well as details about compensation (if any)? 
    \item[] Answer: \answerNA{} 
    \item[] Justification: The paper does not involve crowdsourcing nor research with human subjects.

\item {\bf Institutional review board (IRB) approvals or equivalent for research with human subjects}
    \item[] Question: Does the paper describe potential risks incurred by study participants, whether such risks were disclosed to the subjects, and whether Institutional Review Board (IRB) approvals (or an equivalent approval/review based on the requirements of your country or institution) were obtained?
    \item[] Answer: \answerNA{} 
    \item[] Justification: The paper does not involve crowdsourcing nor research with human subjects.

\item {\bf Declaration of LLM usage}
    \item[] Question: Does the paper describe the usage of LLMs if it is an important, original, or non-standard component of the core methods in this research? Note that if the LLM is used only for writing, editing, or formatting purposes and does \emph{not} impact the core methodology, scientific rigor, or originality of the research, declaration is not required.
    \item[] Answer: \answerYes{} 
    \item[] Justification: LLMs are core components of TriMem, serving as the agent for fact extraction, profile construction, prompt optimization via TextGrad, and final reasoning. The specific models used are listed in Section~\ref{sec:exp_setting}, and all system prompts are provided in Appendix~\ref{app:system prompts}.

\end{enumerate}
\end{document}